\pdfoutput=1
\documentclass[11pt]{article}

\usepackage[final]{acl}

\usepackage{times}
\usepackage{latexsym}
\usepackage{svg}
\usepackage{amsmath}
\usepackage{makecell}

\usepackage[T1]{fontenc}
\usepackage[utf8]{inputenc}

\usepackage{microtype}

\usepackage{inconsolata}

\usepackage{graphicx}
\usepackage{float}
\usepackage{rotating}

\usepackage{booktabs} % For nice table formatting
\usepackage{acro}
\DeclareAcronym{slr}{
  short = SLR,
  long  = Systematic Literature Review
}
\DeclareAcronym{CoT}{
  short = CoT,
  long  = Chain of Thought
}

\usepackage{csquotes}

\usepackage{todonotes}

\definecolor{anne}{rgb}{0,0.5,0.9}
\definecolor{jakob}{rgb}{0.998,0.722,0.635}
\definecolor{instr}{rgb}{0.8,0.67,18}
\definecolor{andi}{rgb}{0.5,0.7,0.18}
\definecolor{chris}{rgb}{0.5,0.25,0.25}

\title{LGAR: Zero-Shot LLM-Guided Neural Ranking\\ for Abstract Screening in Systematic Literature Reviews}

\author{
  Christian Jaumann$^{1,2}$ \hspace{9mm} Andreas Wiedholz$^{1}$ \hspace{9mm} Annemarie Friedrich$^{2}$ \\
  $^{1}$ XITASO GmbH, Germany \hspace{10mm}
  $^{2}$ University of Augsburg, Germany \\
  \texttt{christian.jaumann|annemarie.friedrich@uni-a.de andreas.wiedholz@xitaso.com}
}

\begin{document}
\maketitle

\begin{abstract}
% Motivation
The scientific literature is growing rapidly, % on a daily basis, 
making it hard to keep track of the state-of-the-art.
Systematic literature reviews (SLRs) aim to identify and evaluate all relevant papers on a topic. % literature on a topic. % to synthesize new insights. % to identify the state-of-the-art and to synthesize new insights.
%\wiedholz{we know from experiences that some people believe SLRs are not the way to do it because as soon as they are out, they are outdated again since it takes quite some time. Therefore, sometimes "normal" reviews are preferred. Our automation approach could be beneficial here.}
After retrieving a set of candidate papers, the abstract screening phase %aims to
determines initial relevance. % according to a set of criteria.
% Gap
%Large language models (LLMs) are expected to provide excellent relevance judgment capabilities in this domain, yet,
%Despite being a ranking task in practice \anditodo{would argue with that, at least when humans do it. Make clearer that this is meant for the automation},
To date, abstract screening methods using large language models (LLMs) focus on binary classification settings; existing question-answering (QA) based ranking approaches suffer from error propagation.
%has almost exclusively been addressed as a classification task with LLMs.
%In practice, human post-editing is better supported by providing a meaningfully ranked list.
LLMs offer a unique opportunity to evaluate the SLR's inclusion and exclusion criteria, yet, existing benchmarks do not provide them exhaustively.
% Approach + contributions
We manually extract these criteria as well as research questions for 57 SLRs, mostly in the medical domain, enabling principled comparisons between approaches.
%\annetodo{important: actually we should use them in the Akinseloyin model as well! can we do that? as this is one of our major selling points} % already the case
Moreover, we propose LGAR, a %completely
zero-shot LLM-Guided Abstract Ranker composed of an LLM-based graded relevance scorer and a dense re-ranker.
Our extensive experiments show that LGAR outperforms existing QA-based %abstract ranking
methods by 5-10 pp.~in mean average precision.
Our code and data is publicly available.\footnote{\url{https://github.com/XITASO/lgar/}}

\end{abstract}

\section{Introduction}

The scientific literature grows rapidly, by an estimated average of roughly 7\% per year over the past 22 years \citep{oliveira2022global}, with 3.3 million articles having appeared worldwide in 2022 \citep{Schneider2023Publications}. %\chtodo{fixed citation} %citep{White2019Publications}
Keeping track of the state-of-the-art of a particular research topic is hence inherently difficult.
\acp{slr} are an academic standard %corresponding to a systematic way
of investigating the current state-of-the-art in a specific field.
They follow a strict protocol (e.g., \citet{Kitchenham2013Systematic}). %, in which % several 
%scientific databases 
%, and thus potentially thousands of papers,
%are searched to extract information %about current challenges, trends, etc.,
%for a particular research topic.
Since conducting \acp{slr} is typically very expensive and time-consuming \citep{Michelson2019Significant}, %great
efforts have been made to automate various parts of the process.
Most existing automation approaches focus on facilitating the selection of relevant articles, %studies,
%as this is considered to be
one of the most time-consuming steps % in the \ac{slr} process 
\citep{VanDinter2021Automation}.
Large language models (LLMs) offer unprecedented means and zero-shot capabilities for interpreting text in complex retrieval and ranking scenarios such as abstract screening, where a set of inclusion and exclusion criteria and research questions have to be compared to an abstract.
Recent studies suggest that methods based on LLMs can reduce up to 60-70\% of the workload necessary for paper selection  in the abstract screening phase of SLRs \citep{Tran2023SensitivitySA,Sandner2024ScreeningAF}.

\begin{figure}[t]
\centering
 \includegraphics[width=0.95\columnwidth]{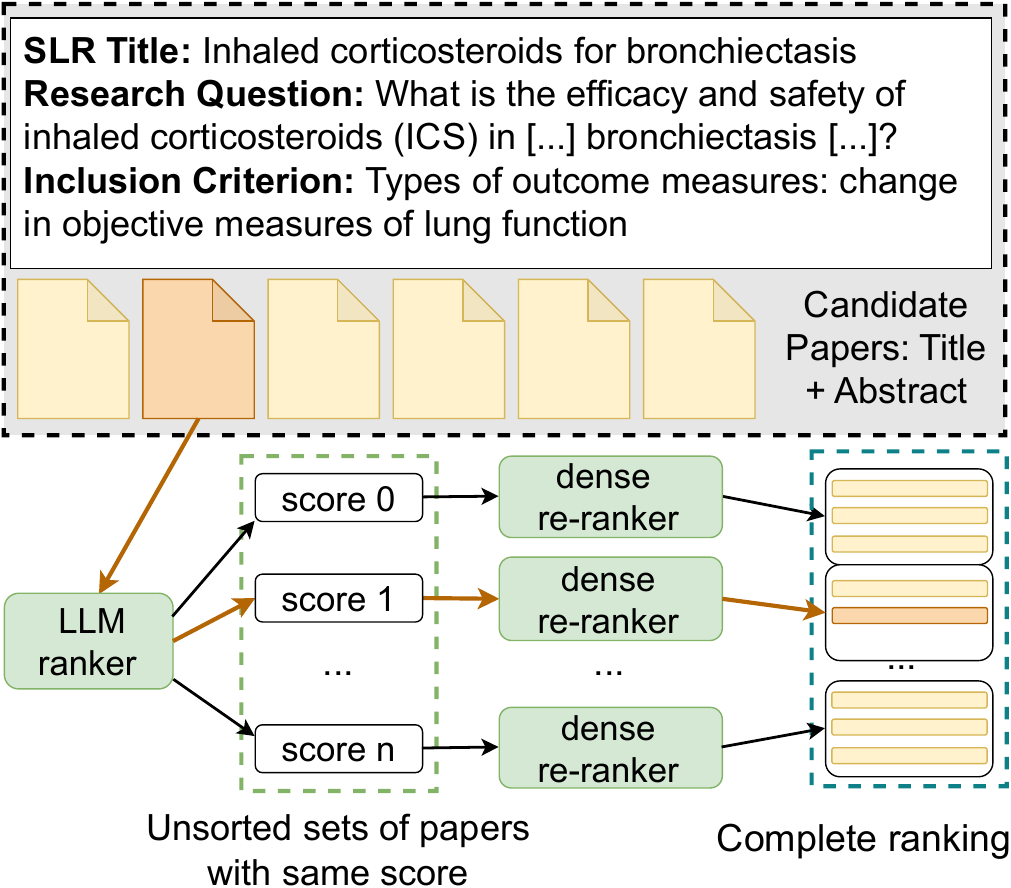}
 % note to myself: add in criteria + research questions, LM ranker = dense ranker, draw lists -- image should clarify model + task (maybe even include human), concrete example of criterion + toy abstract
 % \vspace{1cm}
  \caption {\textbf{LGAR method for abstract screening} consisting of a \textbf{2-stage ranking approach}: (1) the LLM ranker makes use of a realistic specification of the SLR to assign high-level graded relevance scores, (2) dense re-rankers produce a fully ordered list in scalable time.} % \red{increase font size} \red{AF: source of image? please make available to me} \chris{added the drawio file to the folder images}}
  \label{fig:ranking_approach}
\end{figure}

Due to the difficulty of leveraging LLMs for ranking in a scalable manner \cite{Zhu2023Large}, almost all prior work on abstract screening with LLMs works in a binary classification setting, %almost exclusively model a classification setting,
refraining from inducing a ranking %complete ordering
of candidate papers.
%We argue that
In practice, a ranked list is much more useful, % to the human user,
because the decision of the cut-off point can be left to the human user flexibly. % rather than relying on a decision threshold induced by a classifier.
The only existing LLM-based ranker %abstract ranking approach
%\annetodo{which is the other one you mentioned? \chris{"Accelerating Disease Model Parameter Extraction: An LLM-based Ranking Approach to Select Initial Studies For Literature Review Automation" of Sujau et al.}}
\citep{Akinseloyin2024Question}\footnote{The approach was also recently tested on four SLRs on the impact of climate change on diseases \citet{sujau2024accelerating}.} %uses a question-answering setup,
%first
generates questions for SLR criteria and generates answers for each abstract.
%generating binary answers \chtodo{not binary: they had "positive", "negative", or "neural" as labels}\annetodo{ich verstehe das so, dass sie "INCLUDED" or "EXCLUDED" generieren?}
The answers are converted to numeric scores using a %re-purposed
sentiment classifier, extended with embedding-based similarities, and %then
averaged.
While promising, this pipeline-based approach suffers from error propagation in the question generation and sentiment analysis steps.

%This inspiring baseline suffers from two shortcomings: (a) they do not extract inclusion and exclusion criteria from the SLRs in their benchmark exhaustively, (b) it is unclear whether the choice of the sentiment classifier is optimal, and
%(b) experiments are largely carried out with closed LLMs and a temperature of 0.2, complicating reproducibility.
%\annetodo{I am not entire happy about this, think about how to make it stronger}
%\chris{reproducibility is even worse, because they used a temperature of 0.2}

%The inclusion or exclusion of a candidate paper in an SLR depends on a set of criteria and research questions.
Existing abstract screening datasets \citep{DeBruin2023SYNERGY,Kanoulas2019CLEF} provide only part of the SLR criteria. %inclusion and exclusion criteria.
None of them provides the research questions underlying the SLRs.
%\chtodo{\citet{DeBruin2023SYNERGY} added the criteria 2 months ago, but we did not use them since they included references to figures etc.}
%As a result,
Prior ranking studies approximate the information need by using the SLR's title and Boolean document retrieval queries %used during document retrieval
\citep[e.g.,][]{Wang2023Generating,Wang2023Neural}, or use the incomplete set of criteria specified in the SLR abstracts \citep{Akinseloyin2024Question}.
%generally use only the SLR's title as query, which is unrealistic as the SLR's title is not usually available during abstract screening and because it does not specify the criteria.
To evaluate LLM-based rankers
%in a realistic and uniform way across datasets
using the same SLR descriptions as humans do, we manually collect criteria and research questions exhaustively from the full texts of 57 SLRs % of two datasets
and validate our collection process.
This important dataset extension will be made publicly available to foster future work.
%\citet{Akinseloyin2024Question} use a subset of the SLRs' criteria as part of their prompts.

We propose LGAR, a novel %completely
zero-shot \underline{L}LM-\underline{G}uided \underline{A}bstract \underline{R}anker, which directly leverages LLMs to provide graded relevance judgments.
A dense ranker \cite{nogueira-etal-2020-document} then re-ranks the sets of abstracts with the same score.
%Re-ranking of the sets of abstracts having obtained the same score is performed using a dense ranker \cite{nogueira-etal-2020-document}.
% We replicate all relevant baselines using open-weights and open-source models, as well as our complete set of criteria.
Our experiments using our new complete set of criteria and recent open-weights models for all baselines show that %even when provided with the same inputs about the SLR and the abstracts to be ranked,
LGAR outperforms the existing state-of-the-art system \cite{Akinseloyin2024Question} by a large margin (5-10pp. in mean average precision).
%\annetodo{one could argue that the research questions are the main contribution to LGAR vs. Akinseloyin - do we have an ablation? \chris{I'm not entirely sure, what kind of ablation is meant; we have the ablation where LGAR only uses the title not the rqs}}
% --> do we have a clear comparison where the two approaches get the same info on the SLR and one paper? 
% kannst du das hier explizit machen? welche Zeilen der Tabelle? den Leser draufhinweisen mit nem stern oder sowas an der Zeile?
% bei uns gibt es jetzt Mittagessen ;)

%they \red{another shortcoming related to their modeling? experiments?} \chtodo{maybe that they extract the LLMs decision using BART?; furthermore, they did most experiments only with closed source models}

%\anne{TODO describe our method + study}

Our contributions are as follows: 
(1) We provide the exhaustive and validated set of inclusion and exclusion criteria, as well as research questions for two SLR datasets, enabling realistic experiments. %reproducible experiments. % that disentangle the impact of information vs.~prompt style vs.~system architecture.\annetodo{do we really do that in the end?}
(2) We propose LGAR, a zero-shot system for LLM-based abstract screening that performs robustly across datasets and LLM families.
(3) Our extensive experimental evaluation highlights the superior performance of LGAR compared to existing zero-shot and fine-tuned methods. 
% Our detailed ablation study highlights %mportant modeling choices and
% %interesting LLM behaviors such as the extent to which they make use of the suggested judgment scales.
% that for the task of abstract ranking with LLMs, much larger scales with up to 20 options
% % \chtodo{20 options: 0 and 19 are both included on the scale}  % AF: nice catch, I should probably not edit papers that late :/
% are more effective compared to what has been reported for general ranking tasks (up to 5).
We perform a detailed ablation study for the task of abstract ranking with LLMs. We find that using large scales with up to 20 options is most effective. This differs from general ranking tasks, where scales with up to five choices have been reported to be optimal.

\section{Background and Related Work}
%\anne{specify scope of related work here, structure this section a bit!}
This section provides a brief overview of \ac{slr} automation
with a focus on %methods for title and
abstract screening.
%in general, before focusing on approaches that automate the title and abstract screening phase. We summarize the development of different automation strategies, ranging from non-neural approaches to LLM-based solutions. Finally, we provide an overview of LLM-based ranking systems with other applications than abstract screening.

\noindent \textbf{\ac{slr} automation.}
In the first stage of an SLR a set of candidate papers is retrieved from scientific databases, typically using Boolean queries.
The automatic formulation of these queries based on review protocol text has been approached using semantic parsing \cite{alharbi-2018-retrieving,Scells2020Automatic,Scells2020Computational}, rank fusion \cite{scells-2020-teach}, and LLM-based generation \cite{Wang2023Can,Wang2023Generating}. % hence propose to use LLM-generated natural language queries based on Boolean search queries.
SLRs have also been supported by applying summarization methods \cite{Bui2016Extractive} and by extracting fine-grained entity types and relations % information % from medical publications 
\cite{Panayi2023Evaluation}.
%The automation of other stages than abstract screening include
%\annetodo{which? specify}
%defining Boolean queries that are used to gather all the evidence relevant to the review from scientific databases \citep{Scells2020Computational, Scells2020Automatic, Wang2023Can} 
%or extracting data from the selected relevant papers \citep{Bui2016Extractive, Panayi2023Evaluation}.
%\textbf{Active learning tools for SLRs.}
%\annetodo{I would not say whether they are used today, instead, add these citations to the ML techniques above, add info on AL tools as a separate aspect with a focus on explaining the interactive side} 
Most active learning (AL) tools for \ac{slr} automation use traditional machine learning (ML), e.g., AS Review \citep{VanDeSchoot2021Open}, Abstrackr \citep{Wallace2012Deploying}, Colandr \citep{cheng2018using}, FASTREAD \citep{Yu2018Finding}, Rayyan \citep{ouzzani2016rayyan}, or RobotAnalyst \citep{przybyla2018prioritising}. % (all based on traditional machine learning techniques). %, which often not only automate the process of selecting relevant papers with active learning techniques but also facilitate other stages of the SLR workflow.
%In the screening process, the predictions of traditional machine learning models %the model, e.g., \acp{svm} or Naive Bayes,
%are iteratively improved by learning from the inclusion and exclusion decisions of the human reviewers for the data points selected by the model for labeling.
\citet{carvallo-2019-comparing} compare neural classifiers in an AL setup.
%Word2vec \cite{Mikolov2013Distributed} and GloVe \cite{Pennington2014Glove} %\annetodo{add citations!} 
%in an active learning setup.
%In contrast to our approach,
AL assumes substantial training data.
In contrast, we focus on zero-shot settings.
%This interactive process continues until a predetermined stopping criterion is satisfied, such as when the model predicts that none of the remaining abstracts are relevant \citep{VanDeSchoot2021Open}.

\noindent \textbf{Abstract screening.}
%Methods for automatic abstract screening can be broadly divided into binary relevance classification and ranking approaches.
%\wiedholz{sometimes we call it ranking and sometimes priortization approach}\chris{I used both terms as synonyms; is it better to focus on one of them?} \wiedholz{yes, to avoid confusion}
%\annetodo{I agree: Ranking !!}
%The former label each paper as relevant or irrelevant, the latter rank papers according to their relevance to the \ac{slr}.
%\textbf{Non-Neural Automation for Abstract Screening.}
%\annetodo{is this the same as the article selection above? introduce / clarify}
%In general, the literature uses a variety of automation approaches, including various machine learning methods or approaches based on \acp{lm} or \acp{llm}.
%Prior to the advent of \acp{lm} based on the Transformer architecture, common
Early approaches for binary relevance classification for abstract screening apply traditional ML techniques \citep{Kim2014SVM,Khabsa2016Learning,Matwin2010New,Matwin2012Direct}.
% CAMERA READY: can add back details
%, 
%e..g, support vector machines %\acp{svm}  % AF: no need to define acronym if it will not be re-used
%\citep{Kim2014SVM}, random forests \citep{Khabsa2016Learning}, and variants of naive Bayes \citep{Matwin2010New, Matwin2012Direct}. 
%\citep{VanDinter2021Automation, Sundaram2022Automating}. \annetodo{better: add one abstract screening paper that uses the technique after each ML technique as a citation (or multiple)}
%\annetodo{I commented out the discussion because it is a dataset paper rather than a method paper and you give a survey on the methods here. integreate elsewhere, e.g., in the paragraph on datasets if you think it is necessary}
%However, there are still problems with these approaches, such as the fact that the number of papers to be screened is highly dependent on the dataset and can vary widely \citep{Przybyła2018Prioritising, Gates2018TechnologyAssisted}. \wiedholz{don't you have this problem with the other approaches as well? At least the ones the train e.g. LM models} \jaumann{yes, this problem probably occurs in any approach; it was intended to motivate why there is still research conducted in the field despite already having available tools}
%\textbf{\ac{lm}-based Automation for Abstract Screening.}
%The first application of Transformer-based models to abstract screening automation was by
\citet{Qin2021Natural} fine-tune several BERT models \citep{devlin-etal-2019-bert} as binary classifiers and combine their predictions. % in an ensemble.
%Their predictions are then combined as an ensemble. %, which is trained on the predictions of the BERT classifiers.
%\annetodo{? or is this their model? what is their task?}
%\annetodo{recommendation: use firstnameYearFirstWordOfPaper as bibkeys consistently} 
First learning-to-rank setups for abstract screening use non-neural techniques \citep{anagnostou-2017-combining,lagopoulos-2018-learningtorank}.
\citet{lee-2017-study} combines sentence-level relevance scores computed by a convolutional neural network.
\citet{Wang2023Neural} use %an abstract ranking using
a cross-encoder approach following monoBERT \cite{monoBERT}.
%They compare a variety of domain-specific BERT models in zero-shot and fine-tuned settings.
%Furthermore, they compare the effectiveness of two different document representations for the ranking, i.e., title only vs. title + abstract.
%\acp{qlm}\annetodo{is this acronym used later on? maybe also give a high-level term: non-neural retrieval models} and BM25
% While the zero-shot models are outperformed by  non-neural retrieval models, the fine-tuned BERT models demonstrate enhanced performance compared to non-neural methods, including active learning approaches.\annetodo{you can remove the comments on the performance, focus on what they did. performance comparisons  got to experimental section for your chosen baselines}
%\annetodo{use terminology consistently: below, you used citation screening, is this the same as abstract screening?}\jaumann{no, citation screening= abstract screening + full-text screening}
%In a subsequent study, \citet{Wang2023Generating} replace the final title of the \ac{slr} as the ranking query, since the final title might not be available at the citation screening stage. 
 %, utilized for retrieving papers for the \ac{slr}, achieve a performance that was analogous to that of using the final title.
%\wiedholz{Did this then affect the performance? If not, this can be maybe shortened here?} \chris{not much; just thought it might be interesting since we also use the final title -> shows that this could be fixed}
The first abstract screening methods using LLMs all operate in classification settings, applying ChatGPT in a zero-shot setting \citep{Syriani2023Assessing,Guo2023Automated}.
\citet{Robinson2023Bio-SIEVE} also fine-tune Llama \cite{Touvron2023Llama} and Guanaco \cite{dettmers2024qlora} %using instruction tuning
to generate binary inclusion labels. % that they derive from Cochrane.
\citet{Wang2024Zero-Shot} compare a larger set of open-weights LLMs on a calibrated binary classification task. % in a zero-shot setting.

\begin{table*}[t]
\centering
\footnotesize
\setlength{\tabcolsep}{3pt}
\begin{tabular}{lrrlc}
\toprule
\textbf{Dataset}      & \textbf{\# \acp{slr}} &\textbf{ \# Papers} & \textbf{Domain} & \textbf{Avg. inclusion rate (\%) }                                  \\
\midrule
SYNERGY      & 26 & 169288          & medicine (21), psychology (6), computer science (2), & 4.31\\
& & & chemistry (2), biology (3), mathematics (1) & \\
CLEF TAR2019 \\
\hspace{8mm} -- DTA          & 8  &  30521         & medicine (diagnostic technology assessment)                               & 7.05\\
\hspace{8mm} -- Intervention & 20 &  41996         & medicine (clinical intervention trials)                                & 5.49 \\
\hspace{8mm} -- Qualitative  & 2  &  6536         & medicine (qualitative studies)                                             & 1.32\\
\hspace{8mm} -- Prognosis    & 1  & 3367         & medicine (prognosis)                                                      & 5.70\\
\bottomrule
\end{tabular}
\caption{Overview of labeled benchmark datasets for automating systematic literature reviews. Counts for the domains in SYNERGY do not add up to 26 because several SLRs fall under multiple scientific disciplines.} % counts from https://github.com/asreview/synergy-dataset
\label{tab:overview_datasets}
\end{table*}

%\textbf{LLM-based Automation for Abstract Screening.}
\noindent\textbf{Ranking with LLMs.} % AF: intentionally, this includes abstract screening
%LLMs possess remarkable skills when it comes to judging relevance \cite{}, yet, the generative architecture of LLMs is not naturally suited to large-scale ranking task.
Integrating LLMs with ranking approaches is a highly active research area that has just recently emerged \cite{Zhu2023Large}.
LLMs can be integrated as query rewriters, as dense retrievers, as \enquote{readers} to generate natural language answers, or for data augmentation. %\annetodo{ideally, add citations - I'll see what I can do}
We here focus on their application in the ranking stage, where they can be applied using a \textit{pointwise} \cite{Sun2023Instruction}, \textit{pairwise} \cite{Qin2024Large}, or \textit{listwise} approach \cite{Tang2024Found}. %\annetodo{do we have original citations?}
The latter two are inefficient for long candidate lists, which is usually the case in abstract screening.
In pointwise ranking, the generation likelihood of the query given the document can be used as relevance score \citep{Sachan2022Improving},
%can be divided into query generation \citep{Sun2023Instruction}, which uses the generation likelihood of the query given the document as relevance score,
%and relevance generation \cite{Zhu2023Large},
or the LLM can be prompted to generate a relevance score \cite{Liang2023Holistic}. %\annetodo{we should cite one of the first paper doing this instead of the survey here} %\annetodo{citation missing}
%In query generation, %the LLM is prompted to write a query based on a given document.
%the relevance score of a document is taken to be the generation likelihood of the query given the document \citep{Sun2023Instruction}.
%Relevance generation directly asks the LLM to generate a relevance score for each document-query pair which can then be used to rank the documents.
%This score can be binary (relevant/irrelevant), but 
\citet{Zhuang2024Beyond} show that adding intermediate relevance options can improve ranking performance over binary setups.
As abstract screening queries are long and complex, we turn to relevance generation. %a more suitable method.

To the best of our knowledge, the only existing \textit{ranking} approach for abstract screening using LLMs \citep{Akinseloyin2024Question} uses a QA setup.
For each inclusion/exclusion criterion,
%\annetodo{for the replication, are we using the same criteria? \chris{added both options (our criteria vs their criteria)}},
a question is generated using ChatGPT, and the LLM is prompted to answer them for each paper.
Answers are converted to numerical scores based on whether a BART \cite{lewis-etal-2020-bart} model trained on sentiment analysis assigns \textit{Positive} (1), \textit{Neutral} (0.5), or \textit{Negative} (0). % AF: A bit more detail on this one because it is our main baseline. Really weird that they use the sentiment model :/
The final score for each paper is computed by averaging these %question-level
scores with embedding similarities of the questions, the original criteria, and the abstract.
\citet{Akinseloyin2024Question} report that while promising, their pipeline-based approach suffers from error propagation.
%We replicate their model using open components.
Our own system, LGAR, uses a %simpler but
more effective combination of zero-shot LLM-based pointwise relevance generation and monoT5-based re-ranking.

%They create distinct questions comprising the selection criteria with the LLM,
%prompting the LLM for each question and each paper separately, and converting the answers into numerical scores using BART.
%\red{and paper}\annetodo{right?} 
%These scores are then averaged using the cosine similarity between question and abstract, and the final score is calculated by averaging the scores for each question.
%Compared to the results of \citet{Wang2023Neural}, this approach outperforms zero-shot BERT models, but fails to outperform the fine-tuned ones.

%\textbf{Ranking with LLMs.} 
%\annetodo{add this, even if for other domains! in particular the papers that you cite, plus a few more}
%There are three different methods when using LLMs for unsupervised text ranking: pointwise, pairwise, and listwise ranking \citep{Zhu2023Large}. 
%However, due to the large number of candidate documents in the application of abstract screening, pairwise and listwise ranking seem unsuitable due to lower efficiency. 

\section{Dataset Extensions}
In this section, we describe our extension of two abstract screening benchmarks with inclusion/exclusion criteria and research questions to facilitate more realistic evaluation setups.
%\subsection{Abstract Screening Datasets}
%\label{sec:datasets}
%For evaluating our proposed approach,
We extend the SYNERGY dataset \citep{DeBruin2023SYNERGY} and the test set of the CLEF Technological Assisted Review 2019 (TAR2019) dataset (Task 2) \citep{Kanoulas2019CLEF}.\footnote{License SYNERGY: CC0 1.0 Universal; License TAR2019: MIT}
Both datasets are in English and provide the title of each SLR
%\annetodo{in der alten Formulierung ist unklar: auch den Abstract des SLR? oder bezog sich das auf die Papers? \chris{ja, bezog sich auf die papers}}
and a set of papers labeled with binary relevance judgments.\footnote{The SYNERGY dataset has very recently been extended with inclusion and exclusion criteria also collected from full texts, this is concurrent to our contribution. They do not provide research questions.}
%The TAR2019 test set consists of 31 \acp{slr} in the medical domain, which are divided into the following categories: 20 subjects concerning clinical intervention trials (Intervention), 8 subjects pertaining to diagnostic technology assessment (DTA), 2 subjects for qualitative studies (Qualitative), and 1 subject for prognosis. %\annetodo{the info on the subtopics can be moved to the appendix if we need space} 
%For the evaluations we use the categories of the TAR2019 dataset as individual datasets like in related publications \citep{Akinseloyin2024Question, Wang2023Neural, Wang2023Generating}.\annetodo{why not simply report scores for the entire dataset in the tables?}
%Due to our zero-shot approach, which requires no prior training, the TAR2019 training set is not used.\annetodo{why not instead say for all datasets which partitions you USE, not which you do not use} \wiedholz{also another reason here is that we then can better compare us to others right?} \jaumann{not necessarily: Wang et al. have used it for fine-tuning lms; its rather not used because it is only in medical domain -> would not work for synergy}
%The SYNERGY dataset comprises 26 \acp{slr} of the domains medicine, psychology, computer science, chemistry, biology, and mathematics.
\autoref{tab:overview_datasets} gives an overview of the corpus statistics. % of the two datasets.
The inclusion rate specifies the percentage of abstracts included in the SLR out of the original candidate set retrieved using Boolean queries.
The majority of SLRs address topics in medicine. % plural verb is correct here!

\subsection{Extension with Inclusion/Exclusion Criteria and Research Questions}
%As describe in \autoref{sec:datasets}, the SYNERGY and the TAR2019 datasets only provide the SLRs' titles, which underspecify the information need. 
%\chtodo{don't know if we have to mention that: TAR2019 also provides the boolean query used to retrieve the papers, but i guess we can still argue that this is not realisitic enough}
%When conducting an
For SLRs, inclusion and exclusion criteria as well as research questions are specified explicitly as a basis for decision making. %, as well as research questions as the basis for decision making.
%These are the basis for decision making by the human reviewers during abstract screening.
\citet{Akinseloyin2024Question} use the criteria section from the structured abstracts %\annetodo{manually or with LLM?}\chtodo{they copied the part of the abstract about the selection criteria of each SLR and prompted the LLM for every SLR to generate 5 questions based on this information}
(but no research questions) for TAR2019. %but only from the abstract of the respective \acp{slr}
%and without providing any validation. %\chtodo{they only extract the criteria, not the research questions}
However, the full set of criteria and research questions is not always spelled out comprehensively in the SLR's abstract (e.g., it is not the case for SYNERGY). %, which means that their method does not lead to a comprehensive coverage of the information need.
%\annetodo{are the available? we should compare their recall vs. our data, at least in the appendix! we cannot simply state that our data is more comprehensive without showing it}

To provide the models with a realistic evaluation setup, which can be reproduced and re-used by future work, we manually retrieve the inclusion and exclusion criteria and research questions from the full texts of all 57 SLRs of both datasets. %the SYNERGY and the TAR2019 datasets. %, as this information is the basis for decision making for human reviewers when performing abstract screening.
%Additionally, they do not mention any verification process to ensure that the extracted data are valid. \wiedholz{is that too aggressive?}
%Furthermore, since our approach requires the selection criteria and research questions of each \acp{slr}, we manually annotated\annetodo{retrieved the research questions, inclusion/exlusion criteria} \red{all X SLRs of the two} datasets.\annetodo{advertise more: to facilitate a more realistic evaluation setup - this is one of our selling points!}
%To do this,
%We obtain the full text of all SLRs and extract the selection criteria and research questions manually.
If the review itself does not explicitly mention research questions, we infer them from its objectives or extract them from the review protocols which are provided for the TAR2019 dataset. %\annetodo{why "especially"?}
We focus on primary research questions and objectives.
Some SLRs also provide sub-questions and secondary objectives. We do not extract these because they are not typically defined at the start of the abstract screening phase, but more often during the refinement phase of the SLR when several articles have already been checked in greater detail.
%\annetodo{I did not like the original motivation for exclusion, is my suggestion ok? \chris{sounds good for me}}
%which was especially the case for the TAR2019 dataset.
%Some \acp{slr} defined for each research question additional sub research question which we did not consider as necessary information for the LLM. 
%This is also valid for secondary objectives that were present in some of the medical domain. 

%We also extract inclusion and exclusion criteria from the full text and reformulate them as key points. 
%Especially for the \acp{slr} in the medical domain, % AF: we almost have only medical text
In the full text, the criteria are often combined with explanations, definitions, and reasons for specifying a particular criterion.
We only extract the criteria, separating them from surrounding descriptive text, and reformulate them as key points.
% to better understand the respective criterion.
%We do not add these explanations and definitions to the criteria
%We refrained from extracting all of the additional information for each criterion in order to keep the prompt sent to the LLM shorter and more precise.  % AF: bad reason!! benchmark should model human process, not conflate own automatic modeling decisions
%Additionally, reasons why a specific criterion was selected were also not extracted.
In the medical domain, there are sometimes multiple criteria describing the type of studies or type of participants.
We summarize these into one criterion describing all possible types of the respective category.
We ensure that the extracted text does not refer to other material in the review, such as figures or tables. % that are not available to the LLM.
In very rare cases, SLRs also formulate inclusion and exclusion criteria based on metadata such as publication dates which are taken care of during document retrieval.
%These are usually taken care of during document retrieval (see, e.g., the date restrictions in the Boolean queries provided with TAR2019).
%The retrieved document collections can be filtered for them using rule-based methods.
Our set of extracted criteria focuses on semantic information needs.
%\annetodo{are the provided sets of papers already filtered for them? otherwise, we actually SHOULD have extracted them and added this step (we could still do this post-hoc actually if necessary, as the realistic setup is our main selling point)}

%The set of extracted criteria are all such that they require semantic relevance judgments, i.e., meta-data related criteria such as publication dates which can be filtered in a rule-based way are not part of our data collection.

\subsection{Validation}
The extraction of criteria and research questions for all SLRs was performed by the first author of this paper.
%, an undergrad computer science \chris{I do not have a bachelor degree yet (at the earliest probably April)}.
% TODO: add for camera ready: an undergraduate student of computer science (we leave that out now for anonymity reasons)
We validate the extracted information on 15 randomly selected SLRs (5 \acp{slr} from each the SYNERGY, DTA and Intervention dataset).
% CAMERA READY: specify experience of authors
The second author of this paper %with a Master degree in computer science and 2 years of experience in AI research
extracted the criteria and research questions with the same protocol without prior inspection of the data extracted by the first author.
We align the extracted criteria and research questions using the cosine similarities of their BERT embeddings %base
%\citep{devlin-etal-2019-bert} % AF: cited before
and then verify manually whether they indeed correspond to each other.

For the 15 SLRs in our validation study, the first and second author each extracted 23 research questions, which matched exactly.
This high inter-rater agreement stems from the structured nature of medical and computer science SLRs, which often present a section titled \enquote{Objectives} or explicitly state research questions, which allows straightforward extraction.
Incidentally, each author extracted 99 inclusion/exclusion criteria, of which 97 match.
%\annetodo{hatte echt jeder von euch genau gleich viele? \chris{The total number is the same, but author 1 has annotated one criterion more for 2 SLRs and author 2 has annotated one more for 2 other SLRs, so the total number is the same}}
%\chris{With regard to the research questions, the two authors are in complete agreement, and 96\% of the inclusion/exclusion criteria match exactly.}
%\annetodo{are the 96\% only computed over criteria, or were RQs included in the total? \chris{only criteria: 97 criteria match exactly -> 2 * 2 criteria which do not match ==> 97 / (97+4) = 0.96 (at least that is how I think Andi got 96\%); criteria match perfectly}}
%\annetodo{I would calculate: 97/99 match perfectly to get percentages, but let's just report the raw numbers instead! \chris{makes sense}}
Disagreements can be narrowed down to one author including more details to an criterion than the other one or including/neglecting a criterion that was not mentioned explicitly as criterion but described as important factor in the review process.
Overall, we include that inclusion/exclusion criteria and research questions have been extracted reliably for our extension of the SYNERGY and CLEF TAR2019 datasets.

%To evaluate if the two authors agree on the extracted data, the cosine similarity for each data item, i.e. inclusion/exclusion criterion or research question was determined.
%To achieve this, embeddings with the BERT base model (uncased) for the respective items and the similarity between these embeddings were calculated.
%For reproducibility, we provide the raw results from these comparisons in our replication package. \wiedholz{@anne: See data folder in this project}
%We achieved an inter-rater agreement (Cohen's Kappa) of ... .

%, and to exclude criteria that cannot be verified by LLMs, such as the
%requirement for papers to have been published within a certain timeframe are not collected, since the datasets do not include the year of publication of each paper.

% =============================================================================================================

%\section{Two-stage Ranking Approach}\annetodo{can anyone think of a better name for the method?}
\section{LLM-Guided Abstract Ranking}

%The ranking approach employed in this work utilizes
We propose LGAR, a ranking model comprised of a two-stage process to determine the relevance of a screened paper, as illustrated in \autoref{fig:ranking_approach}. 
In the first stage, the LLM assigns each paper a relevance score on a scale from $0$ to $k$, similar to the LLM-based ranking approach \textit{DIRECT(0,k)} of \citet{Guo2024Generating}. %, which they used as a comparison to another ranking approach that generates ranking scores based on a set of criteria from different perspectives. Their results indicate that increasing the rating intervals from (0,10) to (0,20) does not further improve the LLM's ability to differentiate between relevant and non-relevant documents.
%\annetodo{which task did they address and did they also identify a limit?}

Ranking for abstract screening requires producing an ordered list of potentially thousands of papers.
The differentiation that LLMs are able to produce is limited to the range of a dozen relevance scores \citep{Zhuang2024Beyond}, i.e., the first ranking step will produce many ties.
%lly many more papers than different relevance scores, several papers will have the same score.
%To obtain a fully ranked list of papers,
We hence produce the global ranking by additionally ranking the papers within each set of papers that have obtained the same relevance score by the LLM using a zero-shot dense ranking model. % in the second stage of the ranking process.
Our proposed two-stage ranker is modular, providing for modular combinations of LLMs and dense rankers.

%In this two-stage process, the stages and thus the LLM and \ac{lm} used are entirely independent from each other, i.e., different models can be combined in an arbitrary manner.

\subsection{Graded Relevance Judgments by LLM}
In the first ranking stage, the LLM is prompted with the SLR's criteria, as well as research questions (shown in \autoref{fig:um_0s}).
The general instruction encoded in the system message  suggests that a paper should only be considered relevant if all inclusion criteria but none of the exclusion criteria are met.
The method is configurable to use different prompting techniques, but the key elements remain the same.
The system message provides %the necessary context for the conversation, i.e.,
the title and research questions
%a brief topic description\annetodo{what is this topic description? or do you just mean the title, criteria, research questions? \chris{just meant title and RQs of the SLR}}
of the \ac{slr} and the associated ranking task, in addition to a definition of relevance in the context of the \ac{slr}.
The user message contains a more specific task description, instructing the LLM to determine a degree of relevance in the form of assigning a score from a Likert scale using the provided selection criteria, and to answer in a predefined format.
It also provides the title and the abstract of the paper.

\begin{figure}[t]
    \centering
    \scriptsize
    \fbox{\begin{minipage}{0.97\columnwidth}
    \sffamily
    \textbf{System Message:}
        You are a researcher conducting a systematic literature review (SLR) with the title '\{title\}'. The review aims to answer the following research questions: '\{research\_questions\}'\ \ Your task is to decide how relevant the provided paper is to the review, given a list of criteria. A paper is relevant if all inclusion criteria but none of the exclusion criteria are met.

        \textbf{User Message:}
        Task: You will be presented with a paper's title and abstract. Your task is to decide how relevant the given paper is to the review. Return a number for your decision ranging from '\{relevance\_lower\_value\}' to '\{relevance\_upper\_value\}', where '\{relevance\_lower\_value\}' means that you are absolutely sure that the paper should be excluded, where '\{relevance\_upper\_value\}' means that you are absolutely sure that the paper should be included, and where an intermediate value means that you are unsure. Please read the title and the abstract carefully and then make your decision based on the provided inclusion and exclusion criteria.\\Title: `\{title\_paper\}' \hspace{5mm}Abstract: `\{abstract\}'\\Inclusion criteria: `\{inclusion\_criteria\}'\\Exclusion criteria: `\{exclusion\_criteria\}'\\Give your answer in the following format: \\ ```Decision: \{relevance\_lower\_value\} - \{relevance\_upper\_value\}```
    \end{minipage}}
    \caption{Zero-shot prompts used in LGAR.} %\anne{references to appendix: only to appendix section, not to figures}}
    \label{fig:um_0s}
\end{figure}

%\textbf{Generation of few-shot examples.}
\noindent\textbf{Prompting Techniques.}~LLM responses are highly dependent on the type of prompt used \citep{Brown2020Language, wu2023chain, Wang2023Self}.
Therefore, we test several different prompting techniques, i.e., zero-shot, \ac{CoT}, and \ac{CoT} with self-consistency (n=3).
We also tried 2-shot variants without observing any improvements (see Appendix \ref{sec:2-shot}). %, so we stick to the zero-shot setting.
In self-consistency setups, the relevance scores of each paper are averaged over three runs.
For \ac{CoT} prompting, we instruct the model to \enquote{think step by step} \citep{Kojima2022Large}.

\noindent\textbf{Output Parsing.}
To make the responses as deterministic and reproducible as possible, we use a temperature of 0.
The scores are extracted using regular expression %(\texttt{rf"Decision: (\\d+)"})
and we verify that they are in the desired relevance scale. %}\annetodo{one sentence on how you extract them - regex? is it always just a number?}
If score extraction fails, and when using self-consistency, % or when handling responses from which no relevance score could be extracted.
%\wiedholz{Faulty response means that the relevance score cannot be extracted automatically from the response?}
%In these cases,
the temperature is set to 0.5 to encourage more diverse results. % because we want the model to be more creative.
In the rare case that the LLM does not respond in the expected format, the request is repeated up to three times, without providing the model's previous faulty response.
%\annetodo{with the same history or a fresh start with a different temperature?}
If the response is still not in the expected format, the paper is assigned the average of the sum of all LLM generated relevance scores for the respective \ac{slr}. %, since it cannot be known whether the paper is relevant or not.

\subsection{Neural Dense Re-Ranking}
%\annetodo{this is somewhat unclear: what is the query? it can't be the entire prompt - did I rephrase this correctly? \chris{yes, is correct like this}}
In the second stage, we re-rank the papers within the groups generated by the LLM with neural dense retrievers backboned by pretrained language models such as monoBERT \cite{Nogueira2019Passage} or ColBERT \cite{khattab-2020-colbert}.
These cross- and bi-encoders based on BERT %\cite{devlin-etal-2019-bert}
still rank among the state-of-the-art for re-ranking tasks.
Each query-document pair is the concatenation of the query followed by the title and abstract of the paper in question,
%\annetodo{right? \chris{yes}} 
truncated to 512 wordpiece tokens. %\annetodo{not sure if this strictly the limit of T5, hence phrasing it like this}
As query, we use (1) only the SLR's title following 
\citet{Wang2023Neural} or (2) the title and the research questions of the SLR.
%\annetodo{is this new? \chris{yes, since we are the first to extract research questions}}
We use neural re-rankers that were fine-tuned on the general retrieval dataset MS MARCO \citep{Nguyen2016MSMARCO}, but not specifically tuned for abstract screening. % or the particular SLR. 

%For the dense neural ranker, the documents are represented by title and abstract as by \citet{Wang2023Neural}.\annetodo{over-cites Wang, rather: describe the ranking models used}
%However, for the query representation, it is possible to use two different representations similar to the prompt for the LLM ranker: (1) Title-only like \citep{Wang2023Neural} and (2) Title and Research Questions of the \ac{slr}. 

%In case the text input exceeds the input limit of 512 tokens for BERT models, it is truncated.
%\annetodo{have they been mentioned? I guess they will be added to the rel. work? Devlin citation should occur} \jaumann{added them in rel. work}

\section{Experiments}
To compare LGAR to existing approaches, we conduct an extensive experimental evaluation using our newly constructed benchmark.
In this section, we describe our experimental design, including our selection of models, our choice of evaluation metrics, and the baselines.
We then present our experimental results, followed by several ablation studies.

%This section details the experiments we conducted to evaluate our approach.
%First, we motivate the model selection before introducing the evaluation metrics used in the experiments.
%We then explain the overall experimental design and discuss the baselines used for comparison.

\subsection{Model Selection and Experimental Settings}

To enhance the reproducibility of our results, % and due to the high cost of state-of-the-art closed-source models, we have curated 
our model selection of instruction-following LLMs is constrained to open-weights alternatives.
%\annetodo{this said also open source, are there any open source models?}\jaumann{don't know of any} 
We focus on Llama3.3-70B-Instruct \citep{Dubey2024Llama}, which is one of the best performing open-weights models for its size at the time of writing.
To assess the extent of performance disparities among diverse model families and varying model sizes, we conduct comparative analyses with the following models: Llama3.1-8B-Instruct, Qwen2.5-32B-Instruct,\footnote{\href{https://huggingface.co/Qwen/Qwen2.5-32B-Instruct}{https://huggingface.co/Qwen/Qwen2.5-32B-Instruct}} Qwen2.5-72B-Instruct,\footnote{\href{https://huggingface.co/Qwen/Qwen2.5-72B-Instruct}{https://huggingface.co/Qwen/Qwen2.5-72B-Instruct}} %\citep{qwen2.5}\annetodo{if no paper -- huggingface link as footnote}\chtodo{is their official citation from huggingface, which basically references a blog post -> should we just provided the link to the model on huggingface?},
and Mistral-Large-Instruct-2411 (123B).\footnote{\href{https://huggingface.co/mistralai/Mistral-Large-Instruct-2411}{https://huggingface.co/mistralai/Mistral-Large-Instruct-2411}}
We run all models using 16-bit quantization. %\todo{Didn't we at one point discuss to mention the hardware we ran our experiments on? Or is quantization and models enough information -> Appendix}
%\annetodo{citations for Mistral? did you use Mistral or Mixtral? %\chris{this is the model \url{https://huggingface.co/mistralai/Mistral-Large-Instruct-2411}, however I did not find an official citation}}

\begin{table*}
\footnotesize
\centering
\setlength\tabcolsep{2pt}
\begin{tabular}{ll|rrrrrrr|rr}
\toprule
&&& & &&&&& \multicolumn{2}{c}{WSS} \\
\textbf{Dataset} & \textbf{Model} & MAP & TNR@95\% & R@1\% & R@5\% & R@10\% & R@20\% & R@50\% & @95\%& @100\% \\
\midrule
SYNERGY & random reranking & 4.9 & 6.2 & 1.2 & 5.1 & 10.4 & 20.0 & 50.3 & 3.0 & 3.0 \\
 & BM25 (T+R) & 12.5 & 36.3 & 9.0 & 26.4 & 40.0 & 54.4 & 80.3 & 33.6 & 21.8 \\
 & monoT5 (T) & 18.8 & 52.3 & 14.1 & 37.5 & 54.5 & 69.1 & 91.1 & 50.0 & 35.4 \\
 & \citeauthor{Akinseloyin2024Question} (ours) \textsuperscript{$\dagger$}
 & 34.0 & 63.6 & 25.8 & 55.0 & 67.8 & 80.5 & 93.1 & 59.5 & \textbf{50.6} \\
 & LGAR (T, monoT5) \textsuperscript{$\dagger$}
 & 36.8 & 63.8 & 32.7 & 57.7 & 71.0 & 81.2 & 94.3 & 63.6 & 46.5 \\
 & LGAR (T+R, monoT5) & \textbf{40.7} & \textbf{67.0} & 34.1 & \textbf{59.3} & \textbf{72.0} & 81.9 & \textbf{94.4} & \textbf{65.2} & 49.3 \\        
 & \hspace{2mm} + CoT & 38.8 & 62.9 & 33.4 & \textbf{59.3} & 70.3 & 81.4 & 92.8 & 62.5 & 45.8 \\
 & \hspace{2mm} + CoT (n=3) & \textbf{40.7} & 66.4 & \textbf{34.7} & 58.6 & 70.7 & \textbf{82.6} & 93.5 & 64.5 & 50.4 \\
  & LGAR (T+R, random rerank) & 38.9 & 59.8 & 33.3 & 58.4 & 70.1 & 79.7 & 91.4 & 58.6 & 40.9 \\
 \midrule
TAR2019 & random reranking & 6.7 & 9.9 & 1.1 & 5.5 & 10.6 & 20.2 & 52.0 & 6.6 & 7.0 \\
 & BM25 (T+R) & 21.1 & 44.8 & 12.6 & 34.2 & 46.2 & 60.5 & 82.5 & 40.9 & 30.4 \\
 & monoT5\_3B (T) & 30.1 & 63.2 & 17.7 & 44.6 & 59.3 & 73.4 & 94.0 & 58.2 & 50.7 \\
 & \citeauthor{Akinseloyin2024Question} (theirs) & 42.8 & 68.6 & 23.9 & 53.8 & 70.8 & 83.8 & 96.0 & 63.7 & 55.2 \\
 & \citeauthor{Akinseloyin2024Question} (ours) \textsuperscript{$\dagger$}
 & 45.1 & 70.8 & 26.8 & 57.5 & 72.9 & 83.4 & \textbf{96.6} & 66.1 & 53.0 \\
 & LGAR (T, monoT5) \textsuperscript{$\dagger$}
 & 48.4 & 76.4 & 26.0 & 60.9 & 75.7 & 86.8 & 96.2 & 69.9 & \textbf{61.1} \\
 & LGAR (T+R, monoT5) & \textbf{\itshape 50.6} & \textbf{76.5} & 29.3 & \textbf{63.6} & \textbf{76.7} & \textbf{88.3} & 95.8 & \textbf{71.2} & 60.9 \\
 & \hspace{2mm} + CoT & 43.5 & 72.0 & 25.7 & 57.5 & 69.8 & 86.0 & 96.5 & 67.0 & 60.5 \\
 & \hspace{2mm} + CoT (n=3) & 44.8 & 74.9 & 26.1 & 57.9 & 74.2 & 85.7 & 95.8 & 68.8 & 59.4 \\
  & LGAR (T+R, random rerank) & \textbf{50.7} & 64.8 & \textbf{29.7} & 62.1 & 75.1 & 86.0 & 93.7 & 61.4 & 42.9 \\
\bottomrule
\end{tabular}
  \caption{Zero-shot models; recall values are macro-averages over SLRs. All LLM-based models use Llama3.3-70B. T = title of the SLR, R = research questions of the SLR. \citet{Akinseloyin2024Question}: ours = uses our annotated criteria, theirs = uses their criteria. In all combinations, monoT5 uses the title as query and the LLM the scale 0-19. \textsuperscript{$\dagger$} indicates that the approaches get the same information on SLR and paper. % Bold results highlight the best performing model for each metric. %\red{bold-facing}
  % \red{and ChatGPT?}\chris{no, our replication with Llama but with their criteria -> suggests that our annotation is better for most metrics (except for WSS@100\%)}
  %\red{bei allen Recall-Scores und WSS bitte auf nur 1 Nachkommastelle runden (aus den Original-Zahlen um Rundungsfehler zu vermeiden)} \blue{wir könnten die 0-shot Ergebnisse aus \autoref{tab:prompt_avg_res} hier integrieren, wäre dann ja nur die CoT Zeile. lass uns besprechen: da wir für 0-shot keine self consistency nutzen, evtl. besser komplett weglassen? sonst ist es doch unlogisch, warum wir das bei CoT testen? dann wäre es nur eine Zeile mehr in der Haupttabelle}
  }
  \label{tab:main_res}
\end{table*}

%\blue{Following prior work \cite{Akinseloyin2024Question}, \chtodo{Akinseloyin et al. did not use the SYNERGY dataset}
%\annetodo{add citation one for SYNERGY}
We report test results on the entire SYNERGY dataset (which does not provide any training data), and on the test portion of the TAR2019 dataset.\footnote{In alignment with the general evaluation protocol used in the CLEF TAR 2019 Shared Task, we only use the test split of the CLEF TAR dataset comprising 31 SLRs for evaluating our zero-shot approach. We use the train split of 100 additional SLRs to replicate the fine-tuning of BERT-based models for comparing different model types.}
To develop LGAR, we performed prompt-engineering using Llama3.3-70B-Instruct in interactive experiments using SLRs taken from an additional dataset in the medical domain
%\annetodo{right?}\chris{yes}  % 11 SLRs
\cite{Guo2023Automated} (see Appendix \ref{sec:tuning} in particular for the exploration of different scale sizes) and one unpublished SLR. %\todo{Hint that this might change during time of review?}
%\annetodo{make reference to appendix specific to section}
We also run the zero-shot dense rankers on our tuning data, identifying monoT5 \citep{nogueira-etal-2020-document} in the 3B variant as most effective (also see  Appendix \ref{sec:tuning}).
%\annetodo{do you happen to have the dense ranking results for the Guo dataset as well? if yes would be great to add them to the appendix \chris{not yet, but I could compute them in a few hours, if we want to have them}}
In our ablation tests, we additionally evaluate monoBERT (large) \citep{Nogueira2019Passage} and ColBERTv2 \citep{santhanam-etal-2022-colbertv2}.

%\annetodo{I mention them here because they have a particular architecture, for the LLMs, it is quite interchangeable so I moved them to the experimental section. we probably have to repeat the set of LMs there}
%These models are selected, since they are
%, thereby ensuring their suitability for text ranking tasks.
%Furthermore, this selection covers different Transformer architectures, as well as \acp{lm} of different sizes.
%Our approach uses \acp{lm} exclusively in a zero-shot setting, as the availability of labeled training data for a particular topic in a real-world application scenario is rather implausible.

\subsection{Evaluation Metrics}
\label{sec:metrics}
We adopt the evaluation metrics of the CLEF TAR 2019 challenge \cite{Kanoulas2019CLEF}:
%\footnote{To compute MAP and WSS, we use the evaluation script provided by CLEF TAR 2018.}\chtodo{technically, we are still using their evaluation script, but instead of letting it combine the results, we compute the mean by ourselves (as macro recall)} % AF: dann dokumentieren wir das im SourceCode Readme
%To make our results comparable with the results of the CLEF TAR 2019 Task 2 and other related publications \citep{Wang2023Neural, Wang2023Generating, Akinseloyin2024Question}, we employ the same evaluation metrics. 
%These metrics include
mean average precision (MAP), recall at $k\%$ of top-ranked abstracts ($R@k\%, k=1, 5, 10, 20, 50$), and work saved over sampling (WSS) at different recall levels ($WSS@r$\%, $r=95, 100$),
%\footnote{We do not use the rank of the last relevant document (L\_Rel), since averaging this absolute position across several \acp{slr} of different sizes appears to lack significance.} % AF: we can respond that if a reviewer asks about it
defined as follows by \citet{Cohen2006Reducing}:\vspace{-2mm}

\begin{equation}
\text{WSS@r\%} = \frac{\text{TN} + \text{FN}}{\text{N}} - (1-r)
\end{equation}

%\anne{define WSS here, it is not commonly known}
WSS aims to estimate the reduction in human screening workload for a given recall level. 
However, it has some drawbacks, such as that its value range depends on the inclusion rate of the respective \ac{slr}, making comparisons across multiple \acp{slr} unfair \citep{Kusa2023Analysis}.
Therefore, we additionally
%\annetodo{wouldn't it make sense to list them next to each other in the table? or is the order atm intended to put less emphasis on WSS? \chris{yes, it was the intention to put less emphasis on WSS (only provide it for completeness)}}
provide the min-max normalized version of WSS proposed by \citet{Kusa2023Analysis}, which mitigates these issues and
%and corresponds
reports the True Negative Rate at %a given
recall level $r$:\vspace{-6mm}

\begin{equation}
\text{nWSS@r\%} = \text{TNR@r\%} = \frac{\text{TN}}{\text{TN} + \text{FP}}
\end{equation}\vspace{-4mm}

% For our evaluation, $r=95\%$ is chosen because it is close to total recall and is commonly used as a target in other approaches automating abstract screening \citep{Wang2023Neural, Wang2023Generating, Melo2022FewshotAF, Faria2023Automated, Akinseloyin2024Question, VanDeSchoot2021Open, Yu2018Finding}. 

To compute results, we use the evaluation scripts provided in CLEF TAR 2018 with one adaptation: we found that they compute the average recall of multiple \acp{slr} using micro-averaging:\vspace{-2mm}

\begin{equation}
R@k\% = \frac{\sum_{\substack{\{\text{SLRs}\}}} \text{TP}}{\sum_{\substack{\{\text{SLRs}\}}} (\text{TP} + \text{FN})}
\end{equation}

This gives more weight to larger \acp{slr} or \acp{slr} with a higher inclusion rate.
To obtain a more realistic estimate of performance for an unseen SLR, we instead follow \citet{Akinseloyin2024Question} by first computing recall scores per SLR, and then macro-averaging over the obtained scores.
%do not compute $R@k\%$ by averaging the recall values of \acp{slr}, i.e., computing the macro recall that treats each \acp{slr} as a separate class. Instead, they 
%Since , we do not follow this calculation and compute the recall as macro recall like \citet{Akinseloyin2024Question}.
%\annetodo{the min-max normalized version of WSS is in there? if yes we could simplify the text above \chris{no, it isn't; only MAP and WSS}}
%The results of all \acp{slr} in each category are averaged.

%\annetodo{is each LLM run once, or with different temperatures and averaged? -- ah I think the answer is below \jaumann{yes, only once}}

% \jaumann{TODO: discuss if we really want to use all these metrics:}
% \begin{itemize}
%     \item which recall cutoffs (might be too many currently)?
%     \item which recall calculation? clef tar computes average recall differently
%     %\item leave out L\_Rel? (don't see the point in the metric, when it uses absolute values and tries to get 100\% recall)\anne{yes, I thought we decided that in the last meeting}
% \end{itemize}

%\subsection{Experimental settings}
%\label{ssec:experiment_design}

%\blue{To improve the performance of the approaches, we systematically analyze the influence of different factors, such as the prompting technique, the model (family and size), the scale for the relevance values, or providing the model with different amounts of information (Title vs. Title + RQ), on the performance.}\annetodo{move (note to myself)}

\subsection{Baselines}
As non-neural information retrieval baseline, we provide the results of Okapi \textbf{BM25} \citep{RobertsonZ09}.
%\anne{In the table, the input is title + research question as query, is that correct? because it does not have a length limit to the query?}\chris{yes, exactly}
We also report the results of \textbf{monoT5}, %\cite{Nogueira2019Passage},
which is similar to the zero-shot model of \citet{Wang2023Neural} that uses BERT-base.

The closest existing work to ours is that of \citet{Akinseloyin2024Question}, who use a QA setup and (except for a small case study) ChatGPT.
%\annetodo{is that correct? I only skimmed the paper \chris{yes, they primarily used GPT-3.5}}
For a fair comparison, we replicate the \textit{GPT\_QA\_Soft\_Both\_ReRank} variant of their model (which also uses selection criteria for each SLR) %\annetodo{are they different from ours?} - answer: replaced with ours in our experiments
using Llama3.3-70B-Instruct and the embeddings proposed by \citet{wang2024improving-text} (instead of the closed-source embeddings they use). %text-embedding-ada-002 embeddings).
To increase reproducibility, we use a temperature of 0 instead of 0.2 as in the original work.
%\chtodo{maybe mention that we use a temperature of 0 for reproducibility (they used 0.2)}
%For a fair comparison,
%We provide the model with the same complete set of criteria as our models.
%As shown in \autoref{tab:main_res},
Our replication outperforms their scores provided for the TAR2019 dataset on all metrics except WSS (which is a somewhat unreliable metric and only reported for comparison with prior work).
Detailed scores are provided in Appendix \ref{ssec:appendix_eval_lgar}.

\subsection{Main Results and Findings}
\label{ssec::main_results}

\autoref{tab:main_res} compares the effectiveness of LGAR using Llama3.3-70B, relevance scale 0-19, and monoT5 (title-only) as re-ranker with our baselines.
On both the SYNERGY and the TAR2019 datasets, zero-shot monoT5 using only the title as query outperforms BM25 that compares the titles and abstracts of each paper to the title and research questions of the SLR. %\chtodo{added that BM25 has not only abstract for document but also the title of each SLR}
For reasons explained in section \ref{sec:metrics},
%\annetodo{should be section (subsection is not used in ACL for references), please fix},
we focus on TNR instead of WSS when interpreting our results.
Overall, LGAR achieves the strongest performance over all models, outperforming the QA-based model of \citet{Akinseloyin2024Question} by a large margin.

\noindent\textbf{Impact of dense re-ranking.} Our ablation removing monoT5, %as re-ranker,
randomly ranking the papers within each set, shows that in terms of MAP, the difference is small for SYNERGY and negliglible for TAR2019, which is somewhat expected as the LLM-based relevance judgment defines the overall order of the ranking.
Nevertheless, the dense re-ranker contributes to workload savings %to improving the overall ranking
by moving up articles that were incorrectly assigned a low relevance score by the LLM, which is particularly highlighted by the the strong increase of around 7-12\% in TNR.
Similarly, LGAR profits from having access to the SLR's research questions (this difference is more pronounced on SYNERGY).

\noindent\textbf{Comparison to QA-based system.}
%\annetodo{stimmt das überhaupt? LGAR hat immer all Kriterien + research questions (siehe mein Kommentar in der Intro) - sollten wir hier besser noch einmal LGAR testen ohne RQs? \chris{LGAR (T, monoT5) ist LGAR ohne RQss (RQs werden hier also weder für LLM noch für re-ranking benutzt)}}
LGAR (T, monoT5) and the QA model of \citet{Akinseloyin2024Question} (ours) differ with respect to model architecture and prompts, yet both receive exactly the same information about the SLR and the papers.
Even in this setting (without using research questions), LGAR outperforms the QA model, showing that the performance increase is not merely due to additional information in the prompt but due to model architecture.
However, given that the research questions should be available at the stage of abstract screening when conducting an SLR and that adding them appears to enhance the performance of our approach, it seems reasonable to make use of them.

\noindent\textbf{CoT prompting.}
Instructing the model to think step by step does not improve the performance of LGAR. 
Using self-consistency to make up for erroneous runs improves the situation in particular for SYNERGY, but %does not lead to results %that are better overall.
%\annetodo{I did not understand the point about averaging relevance scores, was it important?
%\chris{not that important, was more a detail why self-consistency makes up for erroneous runs (e.g., when a paper is wrongly assigned a low score in the first run, but in the other two runs a high score, the overall relevance is still very high -> averaging the scores corrected the wrong score of the first run (would not be possible if we only had 1 run)}}
%Using this CoT reasoning may cause the LLM to focus on different aspects when analyzing the relevance of the papers, which does not seem to be beneficial in this application.
%Combining CoT with self-consistency does improve overall performance, possibly because averaging the relevance score of each paper over multiple runs corrects for erroneous relevance judgments of individual runs.
%In sum, CoT prompting with self-consistency comes at a significantly higher computational cost and
does not lead to any performance improvements over standard LGAR. Detailed information can be found in Appendix \ref{sec:2-shot}. %\todo{added appendix}

\subsection{Ablation Tests and Analysis}
We now perform an extensive exploration into various settings of LGAR. Detailed scores are provided in Appendix \ref{ssec:appendix_model_selection}.

\subsubsection{Effectiveness of different LLMs}

\begin{table}[t]
\footnotesize
\setlength{\tabcolsep}{3pt}
\centering
\begin{tabular}{ll|rrr}
\toprule
\textbf{Dataset} & \textbf{Model} & MAP & TNR@95\% & R@20\% \\
\midrule
SYNERGY & Llama3.1-8B & 31.9 & 61.3 & 80.9 \\
 & Llama3.3-70B & 40.7 & 67.0 & 81.9 \\
 & Mistral-123B & 38.8 & 65.8 & 82.0 \\
 & Qwen2.5-32B & 38.7 & 63.2 & 78.6 \\
 & Qwen2.5-72B & \textbf{41.3} & \textbf{68.5} & \textbf{83.0} \\
 \midrule
TAR2019 & Llama3.1-8B & 42.3 & 69.6 & 86.1 \\
 & Llama3.3-70B & \textbf{50.6} & \textbf{76.5} & 88.3 \\
 & Mistral-123B & 48.4 & 70.8 & 86.4 \\
 & Qwen2.5-32B & 47.6 & 73.4 & 84.6 \\
 & Qwen2.5-72B & 50.2 & 73.0 & \textbf{89.0} \\
\bottomrule
\end{tabular}
  \caption{Comparison of LLMs. Scale: 0-19. Re-ranker: monoT5 with title-only as query. 0-shot prompting.
  %\red{Bitte Tabelle so kürzen, dass die scores im heading angezeigt werden (rest kommt in appendix) - achtung R@20 aktuell verrutscht}
  }
  \label{tab:llm_exp-short}
\end{table}

\begin{figure}[ht]
    \includegraphics[width=0.48\columnwidth]{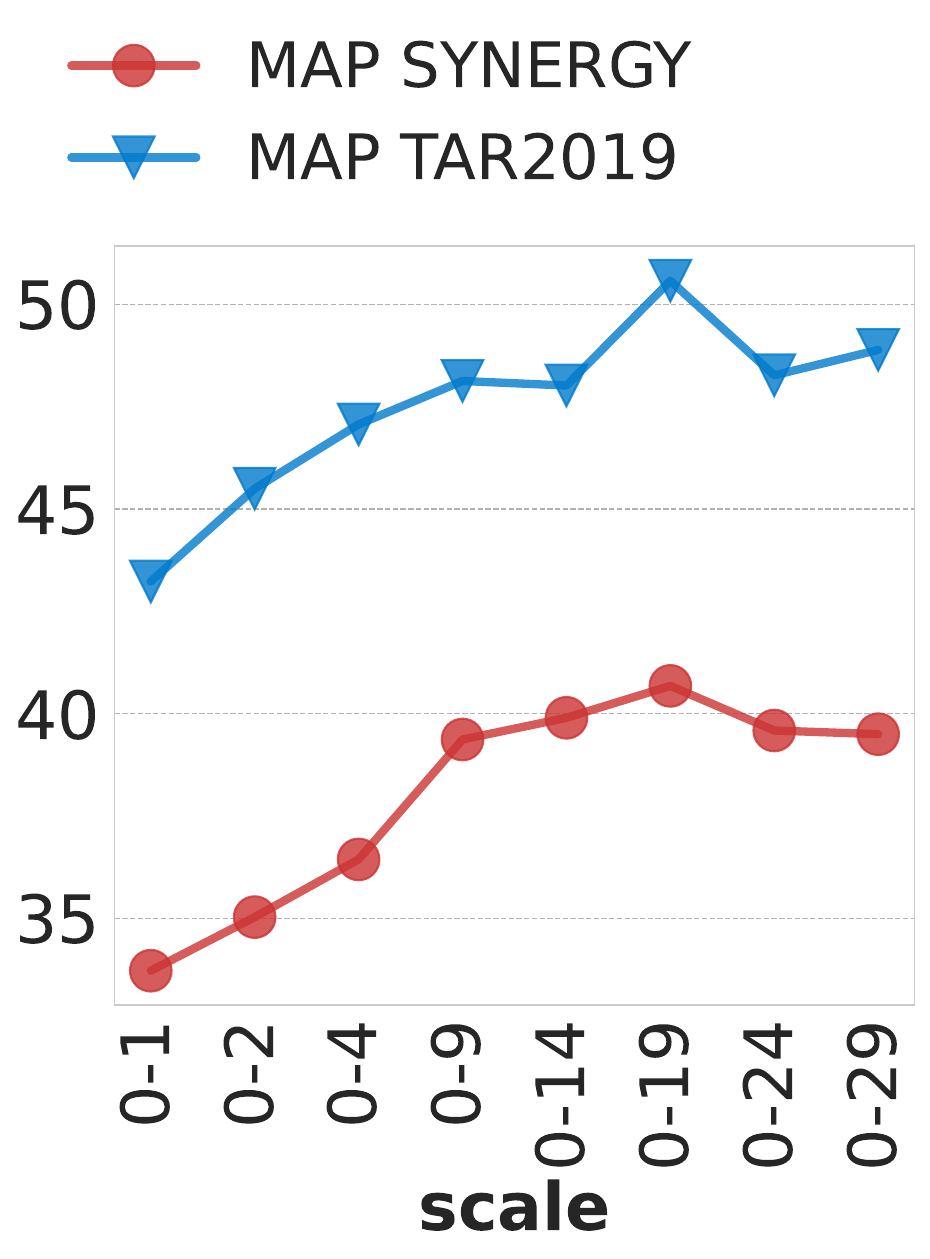}
    \includegraphics[width=0.48\columnwidth]{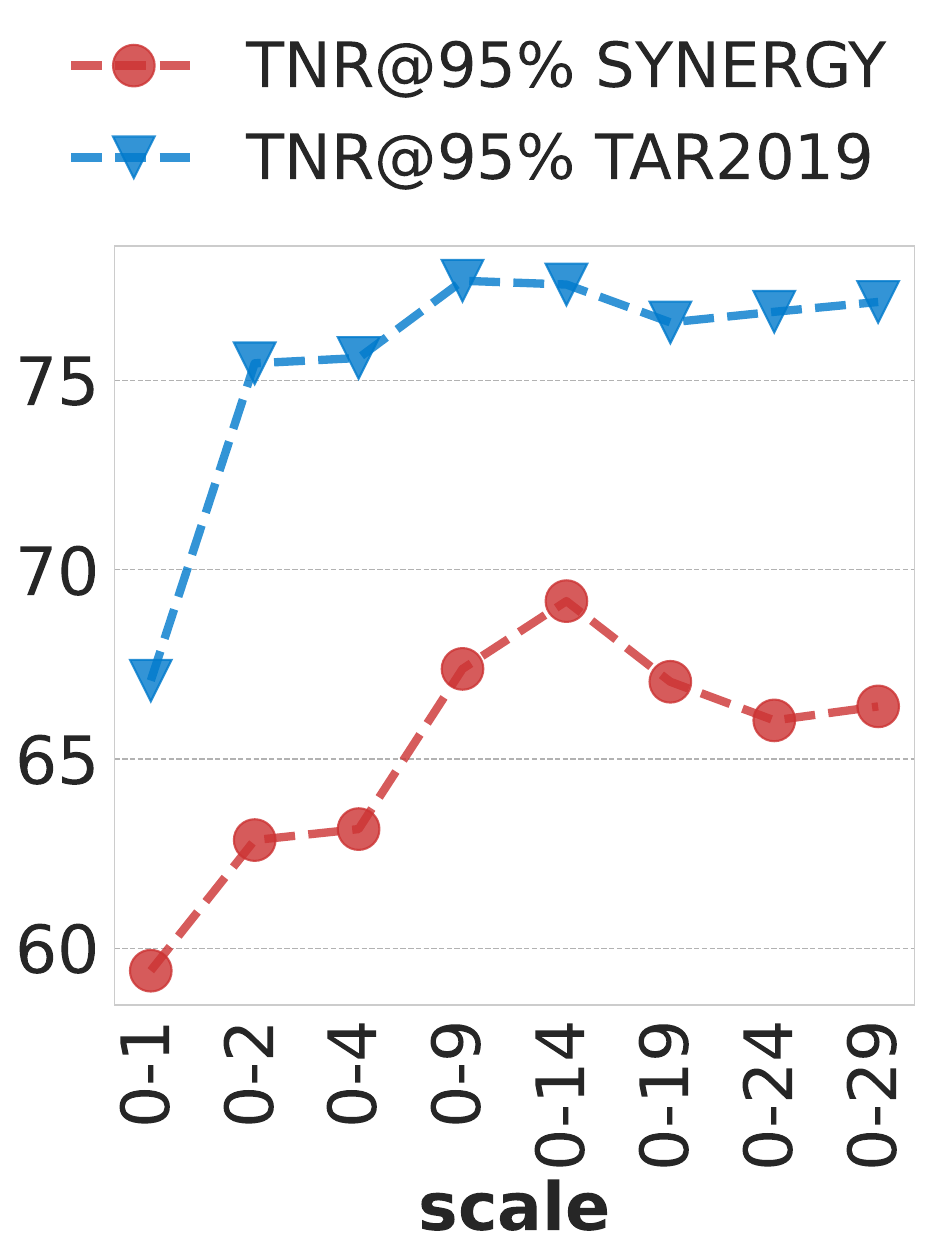}
    \caption{Comparison of scales in terms of MAP and TNR@95\%. The optimum is around 0-14 and 0-19.}%. ideally: tiny barplot see \autoref{tab:scales_avg}, TNR@95
    %blue: MAP, line+triangle=TAR2019, line+circle=SYNERGY,
    %red: TNR, dashed+triangle=... dashed+circle=SYNERGY, alle Kombinationen in Legende, EINE Y-Axis
    \label{fig:scales_avg_plot}
    % AF: ich habe doch zwei plots gemacht, da sie sonst zu flach sind
\end{figure}

\autoref{tab:llm_exp-short} reports the effectiveness of using different LLMs with a relevance scale of 0-19 and monoT5 (title-only) as re-ranker.
Performance degrades for smaller models, but using the larger Mistral-123B did not provide any benefits over Llama3.3-70B.
Qwen2.5-72B excels on SYNERGY and also performs strongly on TAR2019.
Overall, LGAR (which was developed for Llama-3.3-70B) is relatively robust with regard to the underlying LLM (within the same model sizes).
%It shows that using a different model family has little effect on performance, while using smaller models significantly degrades performance.
%Furthermore, it seems that increasing the model size does not necessarily improve the results, at least for the Mistral model family used in this ablation.

\subsubsection{Impact of scale size}
\begin{figure}[t]
  \includegraphics[width=1\columnwidth]{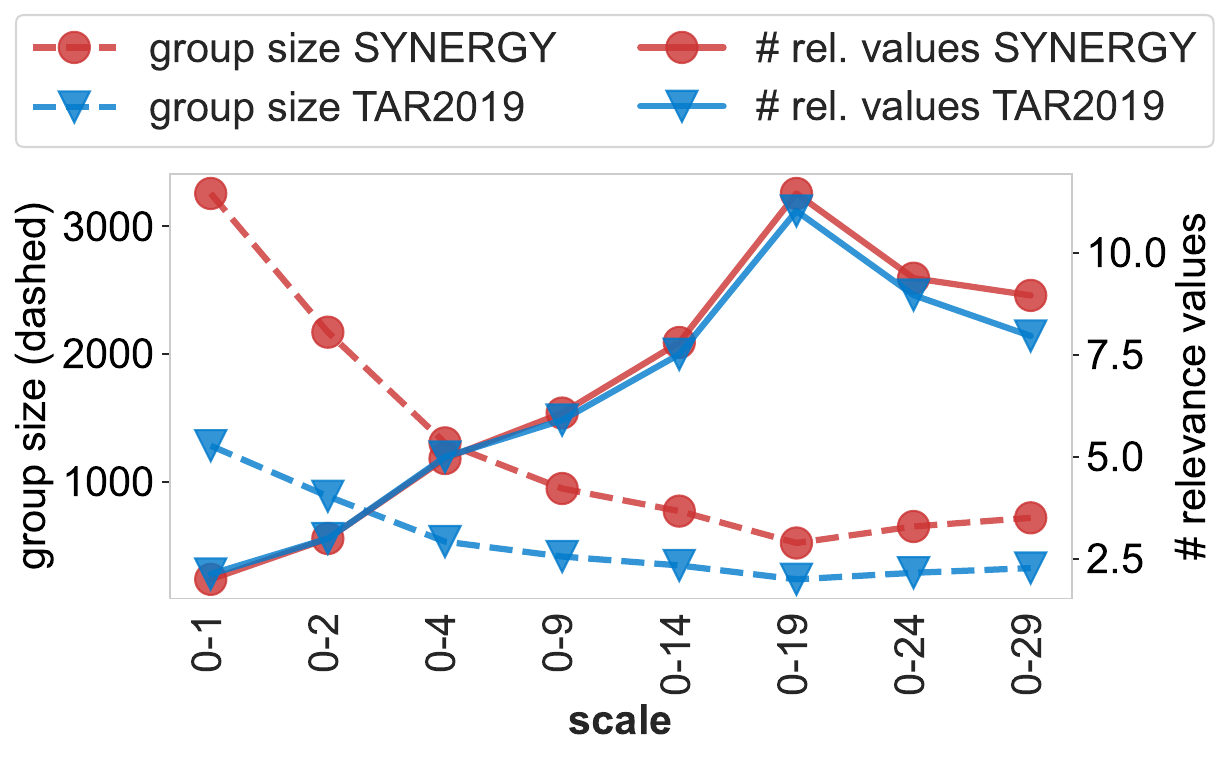}
  \caption {Average number of different relevance scores actually used by LLM and average group size by scale, all SLRs of SNYERGY and TAR2019, %\red{merging all SLRs of both datasets here?} \chris{yes, averaged over all SLRs of both datasets} \red{Increase font size !}
  Llama3.3-70B as LLM ranker and monoT5 (title-only) as re-ranker. 
  %\red{increase font size, Linien, weniger Grid, evtl. nur horizontal+weniger, horinzontal 70\%, legende rechts, avg raus, "# relevance values", text im Bild: lower-case, triangles/circles etc. größer}
  }
  % Anne: I used different colors because now red = always SYNERGY, blue = always TAR
  \label{fig:avg_num_diff_scale}
\end{figure}

%Initially, we chose the scale of 0-19 based on preliminary experimentation with SLRs taken from a third dataset \citep{Guo2023Automated}.
%\annetodo{need to say why we do not evaluate on this one}
%\annetodo{still need to think about whether we report result in the appendix}\citet{Guo2023Automated} \chtodo{if we use the entire Guo set it is larger than the single categories of CLEF TAR (it has about 50.000 documents)}
We here conduct a principled analysis of various choices for scales.
\autoref{fig:scales_avg_plot} reports the performance of LGAR using several different scale sizes.
\citet{Zhuang2024Beyond} found that for general ranking tasks, 0-4 worked best.
By contrast, for our complex ranking task, it is beneficial to increase the scale size even further. 
In terms of workload savings, 0-14 performs best, while the best ranking is produced by 0-19.
Increasing the scale size beyond 0-19 does not further enhance the performance.
Interestingly, this replicates our initial findings on our tuning data, validating our choice of 0-19.
As can be seen in \autoref{fig:avg_num_diff_scale}, %this could be because
the number of different relevance scores actually used by the LLM %actually 
drops for larger scales.
The average number of papers ending up in one group is smallest for 0-19, which means in the 0-19 %\annetodo{inconsistent: (0,19) immer mit 0-19 - should be consistent across the paper, let's not use math mode for this} 
setting, the LLM takes over most ranking workload compared to other settings.
%This suggests that increasing the scale size too much causes the LLM to discriminate less between relevance values.
%At the same time, the average size of groups with the same relevance value does not decrease further, which increases the impact of the \ac{lm} ranker and does not seem to be beneficial for the performance.

\subsubsection{Effectiveness of dense rankers}
\begin{figure}[t]
  \includegraphics[width=1\columnwidth]{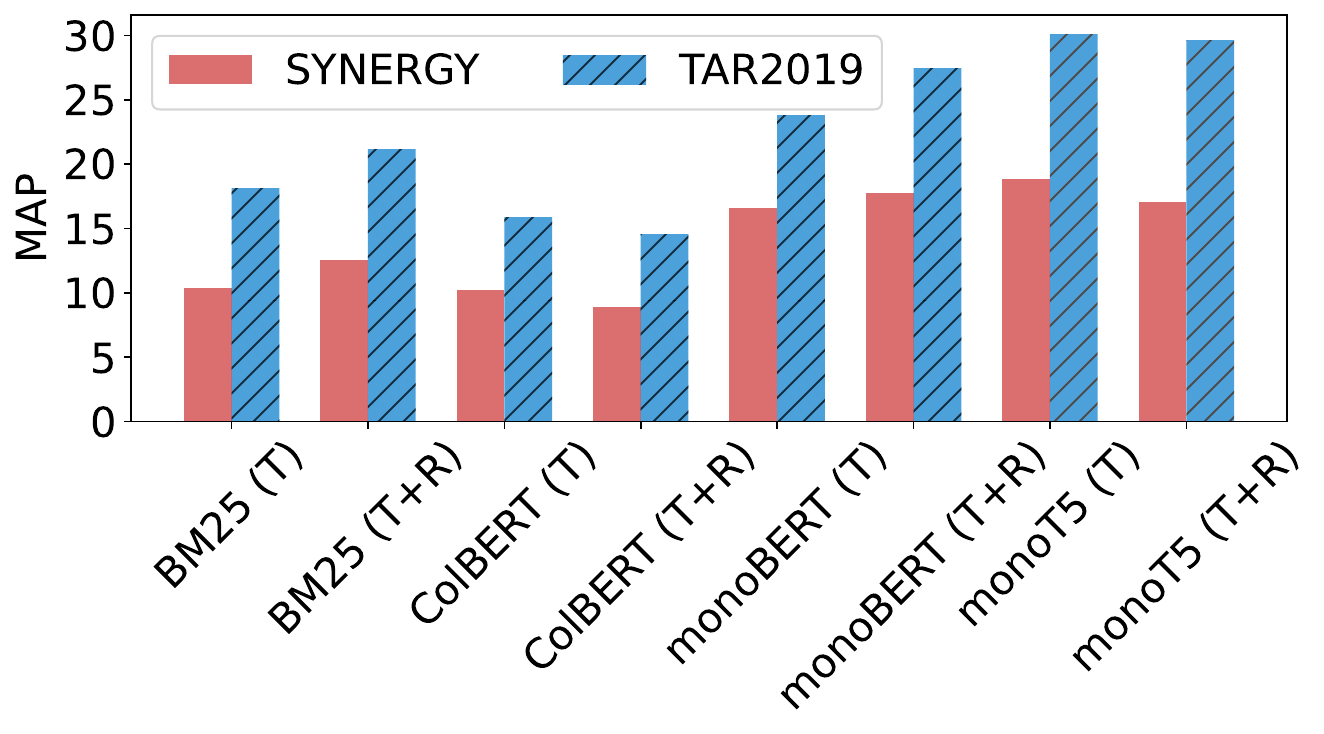}
  \caption {MAP of zero-shot dense rankers (without LGAR). T = title / R = research questions used as query.} %. SYNERGY: normaler Hintergrund, hatched for TAR, bisschen vertikal stauchen und bars etwas breiter, zero-shot} % \red{AF: is this without LLM? then maybe we should just say dense rankers} \chris{yes, is woutout LLM}}
  \label{fig:dense_rankers_ablation}
\end{figure}

In \autoref{tab:ft_model_comparison}, we compare the performance of BM25 with that of three zero-shot dense rankers (without using the LLM step) % trained for text ranking
using two different queries: title-only (T) and title and research questions (T+R).
Details are reported in the Appendix \ref{sec:tuning}.
On both the SYNERGY and the TAR2019 dataset, the best configuration by far uses monoT5 with only the title as query.
% and is therefore used in all subsequent experiments as re-ranker.
While BM25 and monoBERT can deal with the longer input query (T+R), the performance of ColBERT and monoT5 degrades over only using the title.

\begin{table}[t]
    \centering
    \footnotesize
    \setlength{\tabcolsep}{1.5pt}
\begin{tabular}{ll|rrr}
\toprule
\textbf{Dataset} & \textbf{Model} & MAP & TNR@95\% & R@20\% \\
\midrule
Interv. & BioBERT\_ft (T) & 45.9 & 70.8 & 84.2 \\
 & LGAR (T) & 55.0 & \textbf{75.0} & \textbf{88.9} \\
 & LGAR (T+R) & \textbf{56.8} & \textbf{75.0} & 88.4 \\
 \midrule
DTA & BioBERT\_ft (T) & 35.4 & 76.0 & 82.5 \\
 & LGAR (T) & 34.5 & \textbf{78.1} & 84.2 \\
 & LGAR (T+R) & \textbf{38.1} & 76.8 & \textbf{85.0} \\
\bottomrule
    \end{tabular}
    \caption{Comparison to fine-tuned ranker. BioBERT\_ft is our replication of \citet{Wang2023Neural}. Re-ranker for LGAR: monoT5.} %: for Intervention and DTA, make a tiny table with the relevant info (LGAR T+R), BioBERT, - the Wang ft is maybe using a different metric, right? can we compare it if we only list MAP? -- discuss in meeting as well!}
    \label{tab:ft_model_comparison}
\end{table}
%Adding the research questions to the query seems to have a varying impact on the performance of the models. 
%This indicates that for some \acp{lm} the additional information might be more distracting while it improves the ranking of other models.
%The variability in performance across the various categories of datasets may be attributed to the extent of representation of these categories in the training data of the \ac{lm}.

\subsubsection{Comparison to fine-tuned model}
To put LGAR's strong zero-shot performance into perspective, we compare it to the fine-tuned model of \citet{Wang2023Neural}.
We replicate the model to %be able to
report all metrics in the exact same setting. % as for LGAR.
Our replicated model slightly outperforms the results reported by \citet{Wang2023Neural} (see Appendix \ref{ssec:appendix_eval_lgar}). % \todo{appendix added}
We fine-tune BioBERT (large) \cite{Lee2020BioBERT} using the framework of \citet{Gao2021Rethink} and the training splits for the Intervention and DTA
%\annetodo{manchmal upper case - make consistent across paper} 
subset of TAR2019 (there is no training data for the rest of TAR2019 and for SYNERGY).
Our zero-shot LGAR model outperforms the fine-tuned model by 7-10pp.~in terms of MAP (see \autoref{tab:ft_model_comparison}). %\annetodo{Akinseloyin hatten auch schon so was Ähnliches herausgefunden, oder? \chris{nein, die hatten sich nur mit den zero-shot Modellen verglichen - die feintrainierten Modell von Wang et al. waren nämlich leicht besser als Akinseloyins Ansatz}}

%As can be seen in \autoref{tab:main_res_appendix}, our best performing \ac{lm}-based ranker outperforms the zero-shot BERT ranker of \citet{Wang2023Neural}, but cannot compete with a \ac{lm} fine-tuned for this task and domain.
%However, it is noteworthy that our \ac{lm}-based ranker does not necessitate any fine-tuning and exhibits robust performance across diverse domains, as evidenced by the synergy results.
%The significant performance improvement over the previous zero-shot \ac{lm} ranker can be attributed to the significantly larger model size and monoT5's training for text ranking.

\subsubsection{Performance distribution across SLRs}
\begin{figure}[t]
  \includegraphics[width=1\columnwidth]{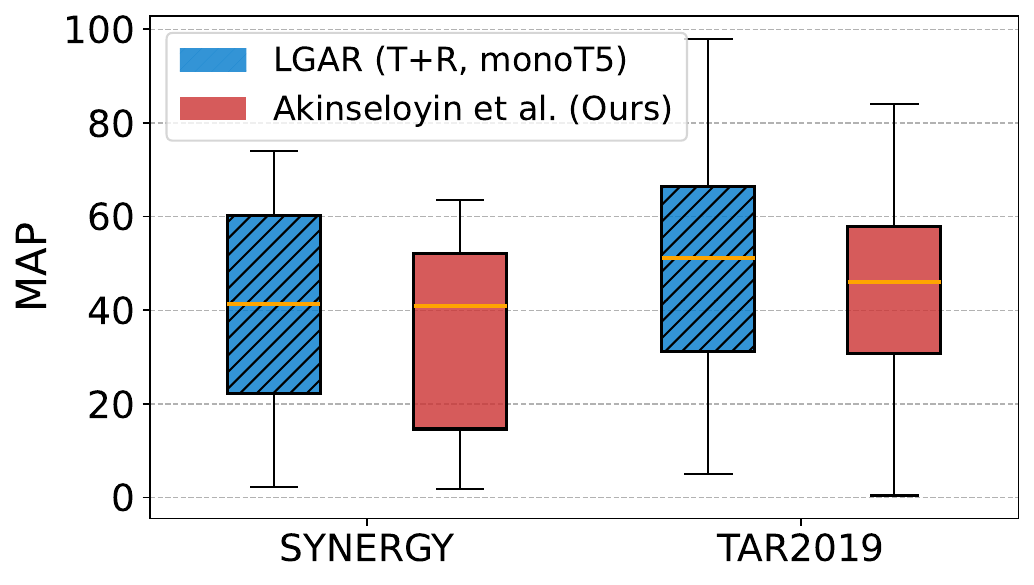}
  \caption {Boxplots of MAP scores of LGAR (T+R, monoT5) and \citet{Akinseloyin2024Question} (ours) on both datasets.
  }
  \label{fig:map_boxplot}
\end{figure}

\autoref{fig:map_boxplot} compares the distribution of MAP scores achieved by LGAR with that by our replication of the approach of \citet{Akinseloyin2024Question} on both datasets. MAP scores differ greatly between SLRs, which is due to their very different inclusion rates.
LGAR appears to have an even greater range of scores due to its higher peak MAP scores. 
Overall, LGAR achieves at least the same performance as the QA-based approach, with a clear tendency to be better.

\section{Discussion and Outlook}
%\annetodo{eher nicht selbstkritisch, sondern mehr Fokus auf "coolen insights"}

%\blue{Christian - hast du Lust einen ersten Draft für das hier zu schreiben? "What did we learn?"}\chtodo{soll das schon selbstkritsch sein oder erst in den limitations? und schreibt man da schon rein, was man noch ausprobieren könnte?}

Our results suggest that LLMs are very effective in interpreting inclusion and exclusion criteria for abstract screening. %and that they can
%$for distinguishing between relevant and irrelevant papers.
They provide a good initial categorization. % that can be used by a dense re-ranker. %to re-rank groups with the same relevance score.
Re-ranking is particularly effective for finding relevant papers accidentally categorized as irrelevant by the LLM, %the \enquote{last} relevant papers,
thus significantly contributing to workload savings.
%For more information refer to \autoref{ssec:appendix_model_selection}. \todo{appendix added}

Compared to the approach of \citet{Akinseloyin2024Question}, it does not seem necessary to split the criteria into several parts, which significantly reduces the number of LLM requests. % required, as we only need one request per paper. %Furthermore, using a sentiment scoring model to extract the LLM decision does not seem necessary for ranking. %, as our direct scoring prompts with a finer grained relevance scale work best.
At the same time, LGAR is less \enquote{explainable} than models evaluating every criterion separately.
We hypothesize that LGAR performs more robustly precisely because it is not forced to evaluate each criterion separately, which likely introduces many noisy scores in the QA-based model.
For many criteria, the abstract may not contain sufficient information to allow a meaningful assessment.
The QA-based baseline is supposed to generate \textit{Neutral} in such cases, yet, detecting when information is insufficient to answer a question is yet a difficult task for LLMs \citep{Ji2022Answer, Gautam2023a}. 
In future work, it would be interesting to further support the human user by identifying the criteria that led to including or excluding a paper.
%Yet, while models cannot robustly detect cases where they are unable to make a meaningful decision, we argue that a better ranking is more important to decreasing human workload for abstract screening.
%As in AL setups, this would also require detecting cases where the LLM is not able to make a meaningful decision and deferring them to the human user.

% Unlike previous approaches \cite{Akinseloyin2024Question, Wang2023Neural, Wang2023Generating}, LGAR has also been evaluated on domains other than medicine.\annetodo{a bit tricky: they might request per-domain results, which we want to avoid (there is very little non-medicine data)}

%Taken together, this results in a significant improvement over the current state-of-the-art.

%\anne{someone might ask: why not first apply the LM and then the LLM for some kind of re-ranking?}
%\red{what about pre-training data contamination?}

%LLMs are very efficient at interpreting the criteria - score combination via sentiment scorer does not seem necessary, our direct scoring prompt works best

%LGAR requires less LLM calls than Akins - should be more efficient

\section{Conclusion}
In this paper, we have proposed LGAR, a novel LLM-Guided Abstract Ranker for SLRs.
%\red{really short, 5-6 lines}
LGAR is based entirely on zero-shot open-weights models, yet outperforms models specifically tuned for this task.
We contribute a dataset extension of exhaustively extracted criteria and research questions and compare to %partially replicated
baselines in a fair way, using the same backbones and inputs, finding that LGAR performs best on two abstract screening datasets.

\section*{Limitations}

The SYNERGY and TAR2019 datasets have been created based on actual SLRs, i.e., the set of relevant papers has been extracted from the SLRs' bibliographies.
Hence, strictly speaking, both datasets that we use in this study provide relevance labels that correspond to the full text screening stage.
Yet, as redistributing the full text of scientific publications is not generally possible due to license restrictions, both datasets only provide titles and abstracts of the papers.
Hence, it is theoretically possible that some papers were still included after the abstract screening stage of an SLR, but excluded in the full text screening stage.
In these cases, the title and abstract of the paper, which are the basis for abstract screening, may not contain the necessary information.
Despite this minor approximation, the SYNERGY and the TAR2019 datasets nevertheless provide suitable benchmarks for automatic abstract screening methods, an assumption that we share with prior work \cite{Akinseloyin2024Question, Wang2023Neural, Wang2023Generating, Wang2024Zero-Shot}.

Another limitation of the evaluation is that even though the evaluation was conducted with 57 different SLRs, the approach could only be tested mainly in the medical domain.
This is due to the composition of the datasets that are widely used when evaluating automation methods for SLRs.
% Even though other domains are represented in the SYNERGY dataset we believe the number of SLRs for these respective domains is too small to give a thorough answer to the question if the performance of the approach is equal to the one in the medical domain.
Although other domains are represented in the SYNERGY dataset, we believe that the number of SLRs for these respective domains is too small to give a thorough answer to the question of whether the performance of the approach is equivalent to that in the medical domain.
%Therefore, we only report the average results on the datasets and not on each of them individually.

% Additionally, since these SLRs were already conducted before 2019 the publications cited in these SLRs could have already been seen by the LLMs in the training. 
In addition, since many of these SLRs were published several years ago, some of their data may have already been exposed to the LLMs in their training.
Hence, we can not exclude the possibility of data contamination.

Furthermore, since LLMs can perpetuate and amplify biases present in the training data, LGAR may inadvertently reinforce these biases when ranking abstracts. This could lead to certain research being unfairly prioritized or overlooked, potentially skewing the outcomes of SLRs.

% Ch: since we discuss few-shot only in the appendix, i would not add this in the limitations here
% As described in \autoref{ssec::main_results} CoT prompting did not improve the performance of LGAR which could be due to multiple reasons.
% In our opinion, one limitation of the experiment design regarding CoT prompting is the generation of few-shot examples.\chtodo{CoT is not necessarily combined with few-shot examples}
% In a SLR the definition of good examples is difficult since a paper can be included or excluded for various reasons.
% This makes it a complex task to generate examples that are representative for all paper that will be screened at this stage of the review.
% Additionally, the examples were generated by a LLM and could not be validated by a domain expert which increases the risk of having wrong reasoning in the inclusion/exclusion of a paper.

\noindent\textbf{Risks.}
% When using this approach in the conduction of a SLR, the main risk is to overlook a paper that is relevant for the research questions that are addressed in the SLR. 
When using this approach in conducting an SLR, the main risk is to overlook a paper that is relevant to the research questions being addressed in the SLR. 
% Overlooking a relevant paper is crucial when writing a SLR which might happen when defining a cutoff in the ranked list when applying LGAR to the abstract screening stage of a SLR.
% We argue that this risk is also present when conducting this manually since manually screening hundreds or thousands of papers is also error prone.
% Usually, to mitigate this risk a SLR has a snowballing stage where the citations of each relevant paper, e.g. after the full-text screening, are screened. 
Missing a relevant paper is problematic when writing an SLR, which can happen when defining a cut-off in the ranked list when applying LGAR to the abstract screening stage of an SLR.
However, we argue that this risk is also present when this is done manually, since manual screening of hundreds or thousands of papers is also error-prone.
%To mitigate this risk, an SLR usually has a snowballing stage where the citations of each relevant paper are screened, e.g., after the full text screening. \chtodo{i do not entirely understand the point: if we would classify a paper as irrelevant, but find it again in snowballing, would we screen it again and change our decision?}

\section*{Acknowledgments}
The authors gratefully acknowledge the scientific support and HPC resources provided by the Erlangen National High Performance Computing Center (NHR@FAU) of the Friedrich-Alexander-Universität Erlangen-Nürnberg (FAU) under the BayernKI project v110ee. BayernKI funding is provided by Bavarian state authorities.

Furthermore, the authors gratefully acknowledge the resources on the LiCCA HPC cluster of the University of Augsburg, co-funded by the Deutsche Forschungsgemeinschaft (DFG, German Research Foundation) – Project-ID 499211671.
% CAMERA READY acknowledge cluster


\clearpage
\appendix

\section{Appendix}
\label{sec:appendix}

\subsection{Initial Interactive Tuning of LGAR}
\label{sec:tuning}

In \textbf{\autoref{tab:guo_scales}}, the results of preliminary experiments are shown. 
These experiments were performed on the dataset provided by \citet{Guo2023Automated}, which consists of 11 medical SLRs with a total of about 50000 papers. 
We excluded the SLR aiha from our evaluation because the provided dataset did not contain any relevant papers.
By not using our test sets, i.e., the SYNERGY and TAR2019 dataset, for the initial optimization of our approach, we aimed to make our results more realistic since we did not optimize our approach on these test sets.
We also refrained from using the train set of the CLEF TAR 2019 dataset for the initial tuning, since optimizing on this dataset, which is quite similar to the corresponding test set, could give our approach an unfair advantage. 
Therefore, we decided to use the dataset of \citet{Guo2023Automated}, which focuses on different medical topics. 
Another reason for using this dataset for the initial tuning was that it is rarely used by other publications, so we wouldn't have any baselines to compare ourselves to on the dataset anyway.

The results of using different scales on this dataset (see \autoref{tab:guo_scales}) suggest that increasing the scale size significantly improves performance up to scale 0-14. The scale 0-19 performs only slightly worse, while further increasing the scale size does not improve performance. Since 0-19 still performs quite well, e.g., in terms of MAP, we decided to use this scale size for our main evaluation, thus giving the LLM as many degrees of freedom as possible without compromising performance.

% The dataset used here \cite{Guo2023Automated} provides the relevance labels after the abstract screening in contrast to SYNERGY and TAR2019 which provide relevance labels after the full text screening. \chtodo{not true, only 2 of them}
% \chris{used it since we did not want to use train set of tar2019 (would maybe optimize for these types of slrs)}
% This dataset is relatively small which is the reason why we do not use it in the work presented mainly in this paper.
% Evaluating the performance on labels after the full text screening puts the performance into a non-ideal context since the model might not have all the information it needs to evaluate the relevance of a paper properly.
% An assumption that we and others \citep{Akinseloyin2024Question, Wang2023Neural} accept due to the lack of datasets providing the relevance labels after the abstract screening.
% Therefore, the results of the model in \autoref{tab:guo_scales} are more realistic.
% The table shows that the scale 0-14 seems to perform better but as for the MAP which we consider one of the most important metrics in this application and the R@50 the results do not differ much and the LLM should get more degrees of freedom, we choose 0-19 as the scale to use which is confirmed later as seen in \autoref{fig:dense_rankers_ablation}.

\begin{table*}[ht]
\footnotesize
\centering
\setlength\tabcolsep{3pt}
\begin{tabular}{ll|rrrrrrr|rr}
\toprule
&&& & &&&&& \multicolumn{2}{c}{WSS} \\
\textbf{Dataset} & \textbf{Scale} & MAP & TNR@95\% & R@1\% & R@5\% & R@10\% & R@20\% & R@50\% & @95\% & @100\% \\
\midrule
Guo & 0-1 & 44.7 & 51.4 & 26.7 & 59.0 & 69.6 & 79.6 & 93.0 & 46.1 & 24.9 \\
 & 0-2 & 47.0 & 71.9 & 26.2 & 60.4 & 74.3 & 89.9 & 97.2 & 65.1 & 33.4 \\
 & 0-4 & 52.2 & 74.8 & 27.7 & 64.2 & 78.5 & \textbf{92.1} & 97.9 & 68.9 & 34.4 \\
 & 0-9 & 51.6 & 75.5 & 28.7 & 62.6 & \textbf{81.4} & 91.7 & 98.0 & 72.2 & 35.4 \\
 & 0-14 & \textbf{54.8} & \textbf{80.9} & 29.5 & \textbf{65.2} & 78.1 & 91.9 & \textbf{99.0} & \textbf{73.0} & 39.1 \\
 & 0-19 & 54.7 & 75.1 & \textbf{30.4} & 63.4 & 80.0 & 91.8 & 98.4 & 69.0 & 37.6 \\
 & 0-24 & 53.8 & 74.8 & 29.0 & 64.0 & 80.4 & 91.8 & 97.9 & 68.0 & \textbf{39.8} \\
 & 0-29 & 51.8 & 76.9 & 28.3 & 61.0 & 79.5 & 91.2 & 98.3 & 70.9 & 39.5 \\
\bottomrule
\end{tabular}
  \caption{Results of different scales on dataset of \citet{Guo2023Automated}. Re-ranker: monoT5 with title-only as query. LLM: Llama3.3-70B.}
  \label{tab:guo_scales}
  
\end{table*}

Similarly, in preliminary experiments when using the dataset presented in \citet{Guo2023Automated}, we also investigate the performance of different dense rankers that are used in the second stage of LGAR (see \autoref{tab:dense_rankers_guo}).
% Similarly, in preliminary experiments when using the dataset presented in \cite{Guo2023Automated} we also investigate the performance of the dense ranker that will be used in the second stage of LGAR.
While monoT5 with title and research questions as query achieves the best ranking performance, using monoT5 with title-only as query significantly improves performance after screening about 20\% of the papers, resulting in a better performance on the metrics regarding potential workload savings. Therefore, we decided to use this configuration as our dense ranker for our main evaluation. The results of these subsequent experiments confirmed our choice to use monoT5 with title-only as the query (see \autoref{tab:lm_exp_avg}).
%results on metrics regarding potential workload savings. Therefore, we decided to use 

% In \textbf{\autoref{tab:dense_rankers_guo}} the results show that the best combination of ranker and information the model uses. 
% monoT5 with title only performs especially good for finding the last relevant papers.
% Here the monoT5 model with title and monoT5 with title and research questions performs best in most metrics which is later also confirmed in our later experiment (see \autoref{tab:lm_exp_avg}).
\begin{table*}
\footnotesize
\centering
\setlength\tabcolsep{3pt}
\begin{tabular}{ll|rrrrrrr|rr}
\toprule
&&& & &&&&& \multicolumn{2}{c}{WSS} \\
\textbf{Dataset} & \textbf{Model} & MAP & TNR@95\% & R@1\% & R@5\% & R@10\% & R@20\% & R@50\% & @95\%& @100\% \\
\midrule
Guo & BM25 (T) & 9.5 & 32.5 & 2.1 & 19.5 & 30.0 & 47.2 & 76.4 & 27.2 & 16.4 \\
 & BM25 (T+R) & 15.5 & 32.2 & 9.8 & 23.8 & 35.7 & 51.5 & 80.6 & 27.0 & 17.4 \\
 & ColBERT (T) & 9.9 & 38.0 & 3.6 & 12.4 & 25.8 & 46.8 & 86.5 & 32.9 & 17.0 \\
 & ColBERT (T+R) & 9.5 & 32.8 & 2.6 & 12.9 & 23.1 & 42.1 & 81.7 & 27.3 & 13.6 \\
 & monoBERT (T) & 14.5 & 41.1 & 7.0 & 28.6 & 42.3 & 61.8 & 86.3 & 37.4 & 20.6 \\
 & monoBERT (T+R) & 18.1 & 41.7 & 7.9 & 34.1 & 46.2 & \textbf{63.5} & 88.4 & 39.6 & 20.7 \\
 & monoT5\_3B (T) & 15.7 & \textbf{45.3} & 8.8 & 34.5 & 47.3 & 63.2 & \textbf{89.8} & \textbf{40.8} & \textbf{21.3} \\
 & monoT5\_3B (T+R) & \textbf{19.1} & 37.5 & \textbf{12.3} & \textbf{37.2} & \textbf{48.6} & 61.8 & 87.0 & 32.8 & 19.7 \\
\bottomrule
\end{tabular}
  \caption{Results of dense rankers on dataset of \citet{Guo2023Automated}. T = title of the SLR, R = research questions of the SLR used as query.}
  \label{tab:dense_rankers_guo}
\end{table*}

\subsection{Detailed Evaluation of LGAR}
\label{ssec:appendix_eval_lgar}

In \textbf{\autoref{tab:main_res_appendix}}, the more detailed results of the main findings presented in this work are shown.
% The TAR2019 test dataset has 4 splits with different topics that are reviewed in the respective SLRs. In the main evaluation we reported the averaged results of these 4 splits in order to keep the table lucid.
% It is worth mentioning that the splits displayed in this table are not equally big, e.g. the intervention split included 20 SLRs in contrast to the prognosis split which only includes 1 SLR.
Here, we report separately the results of all four splits with different medical topics of the TAR2019 dataset.
In the main evaluation, the averaged results of all SLRs of TAR2019 are reported in order to keep the table clear.
It is worth noting that the splits shown in this table are not equal in size, e.g., the Intervention split includes 20 SLRs, while the Prognosis split includes only one SLR.
\begin{table*}
\footnotesize
\centering
\setlength\tabcolsep{2pt}
\begin{tabular}{ll|rrrrrrr|rr}
\toprule
&&& & &&&&& \multicolumn{2}{c}{WSS} \\
\textbf{Dataset} & \textbf{Model} & MAP & TNR@95\% & R@1\% & R@5\% & R@10\% & R@20\% & R@50\% & @95\%& @100\% \\
\midrule
SYNERGY & random reranking & 4.9 & 6.2 & 1.2 & 5.1 & 10.4 & 20.0 & 50.3 & 3.0 & 3.0 \\
 & BM25 (T+R) & 12.5 & 36.3 & 9.0 & 26.4 & 40.0 & 54.4 & 80.3 & 33.6 & 21.8 \\
 & monoT5 (T) & 18.8 & 52.3 & 14.1 & 37.5 & 54.5 & 69.1 & 91.1 & 50.0 & 35.4 \\
 & \citeauthor{Akinseloyin2024Question} (ours)\textsuperscript{$\dagger$} & 34.0 & 63.6 & 25.8 & 55.0 & 67.8 & 80.5 & 93.1 & 59.5 & \textbf{50.6} \\
 & LGAR (T+R, random rerank) & 38.9 & 59.8 & 33.3 & 58.4 & 70.1 & 79.7 & 91.4 & 58.6 & 40.9 \\
 & LGAR (T, monoT5)\textsuperscript{$\dagger$} & 36.8 & 63.8 & 32.7 & 57.7 & 71.0 & 81.2 & 94.3 & 63.6 & 46.5 \\
 & LGAR (T+R, monoT5) & \textbf{40.7} & \textbf{67.0} & 34.1 & \textbf{59.3} & \textbf{72.0} & 81.9 & \textbf{94.4} & \textbf{65.2} & 49.3 \\        
 & + CoT & 38.8 & 62.9 & 33.4 & \textbf{59.3} & 70.3 & 81.4 & 92.8 & 62.5 & 45.8 \\
 & + CoT (n=3) & \textbf{40.7} & 66.4 & \textbf{34.7} & 58.6 & 70.7 & \textbf{82.6} & 93.5 & 64.5 & 50.4 \\
 \hline
Interv. & random reranking & 6.8 & 10.2 & 1.0 & 5.5 & 10.5 & 20.0 & 52.6 & 6.6 & 7.6 \\
 & BM25 (T+R) & 24.9 & 44.6 & 13.6 & 35.9 & 48.1 & 62.3 & 85.2 & 41.0 & 31.7 \\
 & \citet{Wang2023Neural} (0s) & 16.0 & - & 5.4 & 21.0 & 32.8 & 50.4 & - & 36.2 & 33.3 \\
 & monoT5 (T) & 36.0 & 63.7 & 21.7 & 47.5 & 62.6 & 76.8 & 94.0 & 59.0 & 53.8 \\
 & \citet{Akinseloyin2024Question} & 45.0 & - & - & 52.6 & 69.7 & 80.8 & 96.2 & 59.2 & 51.9 \\
 & \citeauthor{Akinseloyin2024Question} (theirs) & 47.1 & 67.1 & 25.3 & 57.4 & 71.1 & 82.8 & 96.3 & 62.7 & 54.2 \\
 & \citeauthor{Akinseloyin2024Question} (ours)\textsuperscript{$\dagger$} & 50.4 & 68.4 & \textbf{30.2} & 60.6 & 73.2 & 83.8 & 96.9 & 65.2 & 49.8 \\
 & LGAR (T+R, random rerank) & \textbf{57.0} & 61.9 & 29.1 & \textbf{65.8} & 78.0 & 87.6 & 94.2 & 60.4 & 45.4 \\
 & LGAR (T, monoT5)\textsuperscript{$\dagger$} & 55.0 & \textbf{75.0} & 27.8 & 62.2 & 76.9 & 88.9 & 96.0 & 68.6 & \textbf{63.2} \\
 & LGAR (T+R, monoT5) & 56.8 & \textbf{75.0} & 28.5 & 65.2 & \textbf{78.6} & 88.4 & 95.6 & \textbf{70.7} & 61.9 \\
 & + CoT & 54.3 & 74.2 & 25.6 & 63.0 & 76.6 & 89.0 & 95.4 & 69.1 & 62.2 \\
 & + CoT (n=3) & 54.5 & 73.8 & 27.0 & 64.8 & 77.6 & \textbf{89.2} & 94.9 & 68.5 & \textbf{63.2} \\
 & \citet{Wang2023Neural} (ft) & 45.6 & - & 21.6 & 58.0 & 73.7 & 84.2 & - & 64.6 & 57.9 \\
 & BioBERT\_ft (T) & 45.9 & 70.8 & 24.5 & 55.5 & 73.3 & 84.2 & \textbf{97.0} & 66.8 & 53.6 \\
 \hline
DTA & random reranking & 7.8 & 5.8 & 1.4 & 5.9 & 10.6 & 20.4 & 49.4 & 3.6 & 2.8 \\
 & BM25 (T+R) & 16.7 & 50.5 & 13.9 & 37.6 & 50.1 & 65.0 & 86.0 & 46.4 & 33.1 \\
 & \citet{Wang2023Neural} (0s) & 9.2 & - & 2.4 & 13.2 & 23.8 & 39.1 & - & 25.8 & 21.0 \\
 & monoT5 (T) & 20.4 & 64.0 & 11.1 & 45.0 & 60.5 & 74.6 & 93.0 & 58.4 & 47.1 \\
 & \citet{Akinseloyin2024Question} & 31.5 & - & - & 43.8 & 59.3 & 76.6 & 94.1 & 55.6 & 50.6 \\
 & \citeauthor{Akinseloyin2024Question} (theirs) & 37.3 & 70.6 & 25.0 & 52.5 & 67.0 & 84.4 & 94.1 & 64.5 & 60.3 \\
 & \citeauthor{Akinseloyin2024Question} (ours)\textsuperscript{$\dagger$} & 35.9 & 77.5 & 23.4 & 56.4 & 74.0 & 85.3 & 94.9 & 70.0 & \textbf{62.6} \\
 & LGAR (T+R, random rerank) & 38.0 & 64.5 & 30.6 & 52.8 & 66.6 & 78.0 & 90.3 & 57.7 & 34.1 \\
 & LGAR (T, monoT5)\textsuperscript{$\dagger$} & 34.5 & 78.1 & 25.8 & 57.6 & 73.1 & 84.2 & 95.2 & 71.3 & 57.0 \\
 & LGAR (T+R, monoT5) & \textbf{38.1} & 76.8 & \textbf{30.8} & \textbf{59.9} & 72.9 & 85.0 & 94.6 & 69.5 & 57.3 \\
 & + CoT & 36.5 & \textbf{79.4} & 20.4 & 57.0 & \textbf{75.4} & \textbf{87.0} & 95.9 & \textbf{71.8} & 59.4 \\
 & + CoT (n=3) & 37.6 & 77.5 & 24.9 & 58.9 & 73.4 & 85.5 & 95.1 & 69.9 & 61.0 \\
 & \citet{Wang2023Neural} (ft) & 31.8 & - & 26.0 & 50.0 & 67.1 & 81.7 & - & 68.6 & 58.4 \\
 & BioBERT\_ft (T) & 35.4 & 76.0 & 25.3 & 54.8 & 69.7 & 82.5 & \textbf{96.7} & 69.5 & 60.4 \\
 \hline
Qualitative & random reranking & 1.6 & 24.6 & 0.9 & 3.4 & 11.0 & 21.7 & 58.0 & 20.0 & 21.6 \\
 & BM25 (T+R) & 6.9 & 38.3 & 3.2 & 16.2 & 26.1 & 36.0 & 46.4 & 33.2 & 18.3 \\
 & monoT5 (T) & 15.0 & 64.2 & 9.9 & 28.4 & 36.0 & 41.0 & 98.6 & 59.4 & 51.1 \\
 & \citet{Akinseloyin2024Question} & 15.9 & - & - & 50.5 & 60.0 & 67.3 & 97.8 & 57.6 & 50.7 \\
 & \citeauthor{Akinseloyin2024Question} (theirs) & 17.8 & 75.8 & 11.3 & 28.4 & 85.1 & 91.9 & 99.6 & 72.0 & 50.3 \\
 & \citeauthor{Akinseloyin2024Question} (ours)\textsuperscript{$\dagger$} & 22.4 & 66.6 & 12.2 & 31.5 & 65.5 & 71.0 & 99.6 & 61.1 & 50.4 \\
 & LGAR (T+R, random rerank) & \textbf{29.9} & \textbf{86.7} & \textbf{39.2} & 62.3 & 74.5 & \textbf{98.4} & 99.3 & \textbf{81.6} & 54.0 \\
 & LGAR (T, monoT5)\textsuperscript{$\dagger$} & 27.3 & 81.0 & 13.5 & 61.5 & 70.5 & 73.6 & \textbf{100.0} & 75.4 & 71.8 \\
 & LGAR (T+R, monoT5) & 29.1 & 86.6 & 38.5 & 62.4 & 69.1 & 98.2 & \textbf{100.0} & 80.6 & \textbf{80.1} \\
 & + CoT & 26.1 & 81.3 & 13.1 & 63.3 & 90.5 & 96.0 & \textbf{100.0} & 78.5 & 75.7 \\
 & + CoT (n=3) & 28.2 & 82.0 & 13.1 & \textbf{63.7} & \textbf{91.4} & 97.3 & \textbf{100.0} & 81.0 & 73.5 \\
 \hline
Prognosis & random reranking & 6.3 & 6.3 & 1.4 & 6.0 & 11.0 & 20.0 & 50.8 & 1.8 & 0.7 \\
 & BM25 (T+R) & 9.2 & 15.3 & 0.5 & 7.8 & 16.2 & 38.0 & 73.4 & 10.2 & 5.1 \\
 & monoT5 (T) & 18.5 & 43.9 & 5.7 & 18.2 & 31.2 & 59.9 & 92.7 & 36.7 & 16.6 \\
 & \citet{Akinseloyin2024Question} & 43.0 & - & - & 40.0 & 65.3 & 80.0 & 98.4 & 54.3 & 28.9 \\
 & \citeauthor{Akinseloyin2024Question} (theirs) & 49.4 & 68.3 & 13.0 & 43.8 & 66.2 & 82.8 & 97.9 & 60.7 & 43.8 \\
 & \citeauthor{Akinseloyin2024Question} (ours)\textsuperscript{$\dagger$} & 57.5 & 71.8 & 14.1 & 54.2 & 74.5 & 85.4 & 99.0 & 63.7 & \textbf{44.6} \\
 & LGAR (T+R, random rerank) & 67.1 & 81.1 & \textbf{15.1} & \textbf{62.1} & 84.4 & 93.3 & \textbf{99.8} & 72.8 & 42.3 \\
 & LGAR (T, monoT5)\textsuperscript{$\dagger$} & 67.7 & 81.2 & \textbf{15.1} & 62.0 & 83.8 & 93.8 & 99.5 & 74.1 & 29.8 \\
 & LGAR (T+R, monoT5) & \textbf{67.9} & 83.6 & \textbf{15.1} & 61.5 & 84.4 & 93.8 & 99.5 & 74.3 & 32.8 \\
 & + CoT & 65.2 & 82.4 & 14.6 & 61.5 & \textbf{84.9} & 93.8 & 99.5 & 73.4 & 34.5 \\
 & + CoT (n=3) & 67.1 & \textbf{84.9} & \textbf{15.1} & 60.4 & \textbf{84.9} & \textbf{95.3} & 99.5 & \textbf{75.5} & 32.6 \\
\bottomrule
\end{tabular}
  \caption{Main results (detailed). T = title of the SLR, R = research questions of the SLR. \citet{Akinseloyin2024Question}: ours = uses our annotated criteria, theirs = uses their criteria. In all combinations, monoT5 uses the title as query and the LLM the scale 0-19. \citet{Wang2023Neural}: 0s=zero-shot, ft=fine-tuned. \textsuperscript{$\dagger$} indicates that the approaches get the same information on SLR and paper.
  %\red{please format to one decimal point, reorder as in main part of the paper}
  }
  \label{tab:main_res_appendix}
\end{table*}

\subsection{Detailed Results of Ablations}
\label{ssec:appendix_model_selection}

% \textbf{\autoref{tab:llm_exp}} shows detailed results of the experiment that determines the best suited LLM for LGAR. 
\textbf{\autoref{tab:llm_exp}} shows detailed results of using different LLMs as initial ranker of LGAR. 
% Llama3.3-70B is performing best across most of the metrics for the TAR2019 dataset which is also the trend when evaluating on the SYNERGY dataset.
Llama3.3-70B performs best on most metrics for the TAR2019 dataset, which is also the trend when evaluated on the SYNERGY dataset.
%For other metrics, Llama3.3-70B performs only slightly worse than the best performing model which leads to the decision to use Llama3.3-70B for all upcoming experiments.

\begin{table*}
\footnotesize
\setlength{\tabcolsep}{3pt}
\centering
\begin{tabular}{ll|rrrrrrr|rr}
\toprule
&&& & &&&&& \multicolumn{2}{c}{WSS} \\
\textbf{Dataset} & \textbf{LLM} & MAP & TNR@95\% & R@1\% & R@5\% & R@10\% & R@20\% & R@50\% & @95\%& @100\% \\
\midrule
SYNERGY & Llama3.1-8B & 31.9 & 61.3 & 22.9 & 53.4 & 67.0 & 80.9 & 93.3 & 60.5 & 45.4 \\
 & Llama3.3-70B & 40.7 & 67.0 & \textbf{34.1} & 59.3 & 72.0 & 81.9 & \textbf{94.4} & \textbf{65.2} & \textbf{49.3} \\
 & Mistral-123B & 38.8 & 65.8 & 33.1 & \textbf{61.6} & \textbf{73.3} & 82.0 & 93.6 & 63.4 & 47.2 \\
 & Qwen2.5-32B & 38.7 & 63.2 & 33.4 & 57.0 & 68.1 & 78.6 & 93.7 & 61.4 & 45.8 \\
 & Qwen2.5-72B & \textbf{41.3} & \textbf{68.5} & 32.7 & 60.0 & 72.0 & \textbf{83.0} & 94.1 & 64.8 & 47.7 \\
 \midrule
TAR2019 & Llama3.1-8B & 42.3 & 69.6 & 24.7 & 57.5 & 71.5 & 86.1 & 95.2 & 65.3 & 55.1 \\
 & Llama3.3-70B & \textbf{50.6} & \textbf{76.5} & 29.3 & \textbf{63.6} & \textbf{76.7} & 88.3 & 95.8 & \textbf{71.2} & \textbf{60.9} \\
 & Mistral-123B & 48.4 & 70.8 & 28.5 & 62.7 & 75.3 & 86.4 & 95.9 & 68.1 & 55.1 \\
 & Qwen2.5-32B & 47.6 & 73.4 & 28.6 & 60.4 & 74.0 & 84.6 & \textbf{96.2} & 68.6 & 58.9 \\
 & Qwen2.5-72B & 50.2 & 73.0 & \textbf{29.8} & 63.1 & 76.1 & \textbf{89.0} & 95.8 & 68.9 & 56.0 \\
\bottomrule
\end{tabular}
  \caption{Comparison of LLMs. Scale: 0-19. Re-ranker: monoT5 with title-only as query. 0-shot prompting. }
  \label{tab:llm_exp}
\end{table*}

%\textbf{\autoref{tab:scales_avg}} shows the experiment that confirms the results from preliminary experiments (see \autoref{tab:guo_scales}). 

\begin{figure}[H]
  \includegraphics[width=1\columnwidth]{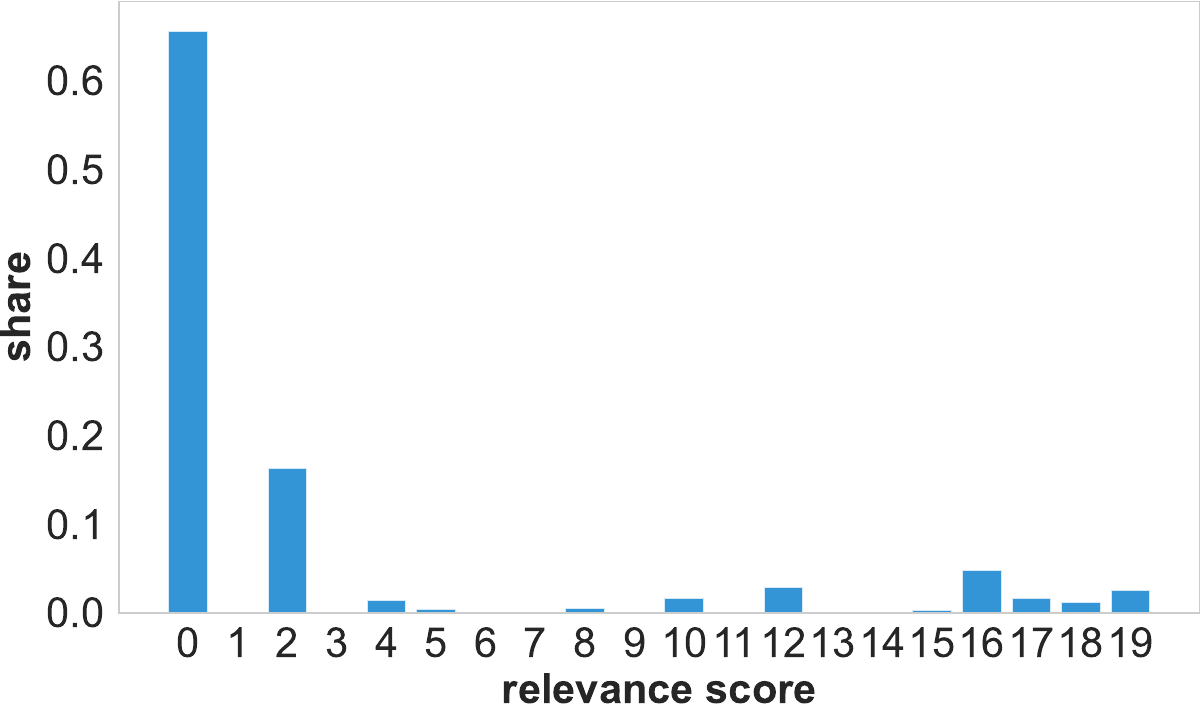}
  \caption {Distribution of relevance scores averaged for all SLRs of SYNERGY and TAR2019. LLM: Llama3.3-70B. Scale: 0-19. Re-ranker: monoT5 with title-only as query.}
  \label{fig:dist_rel_vals}
\end{figure}

\textbf{\autoref{tab:scales_avg}} shows the results of using different scales on the SYNERGY and TAR2019 datasets.
The results confirm our choice to use the 0-19 scale, which was identified as one of the best performing scales in preliminary experiments (see \autoref{tab:guo_scales}). 
%In this experiment, it can be seen more clearly that the 0-19 scale performs best on most of the metrics.
\autoref{fig:dist_rel_vals} depicts the distribution of relevance scores on the relevance scale averaged over all SLRs. %scales. 
Most papers receive a score of 0, which is expected, since most of the papers retrieved as candidates from scientific databases are not relevant to the respective SLR.
%Most papers get a score 0 which is expected since usually most of the papers are not relevant for the SLR at this stage of conduction.

\begin{table*}[ht]
\footnotesize
\centering
\setlength{\tabcolsep}{3pt}
\begin{tabular}{ll|rrrrrrr|rr}
\toprule
&&& & &&&&& \multicolumn{2}{c}{WSS} \\
\textbf{Dataset} & \textbf{Scale} & MAP & TNR@95\% & R@1\% & R@5\% & R@10\% & R@20\% & R@50\% & @95\% & @100\% \\
\midrule
SYNERGY & 0-1 & 33.7 & 59.4 & 29.2 & 55.1 & 67.6 & 79.3 & 93.2 & 58.2 & 41.2 \\
 & 0-2 & 35.0 & 62.9 & 30.8 & 56.0 & 70.1 & 82.1 & 93.6 & 63.6 & 45.3 \\
 & 0-4 & 36.4 & 63.2 & 30.6 & 56.6 & 70.0 & 80.6 & 93.7 & 63.8 & 44.9 \\
 & 0-9 & 39.4 & 67.4 & 32.3 & 57.6 & 69.2 & 82.2 & \textbf{94.6} & 64.8 & 50.1 \\
 & 0-14 & 39.9 & \textbf{69.2} & 33.3 & 57.4 & 69.7 & \textbf{82.3} & 94.5 & 65.5 & \textbf{50.8} \\
 & 0-19 & \textbf{40.7} & 67.0 & \textbf{34.1} & \textbf{59.3} & \textbf{72.0} & 81.9 & 94.4 & 65.2 & 49.3 \\
 & 0-24 & 39.6 & 66.0 & 32.7 & 57.8 & 70.9 & 81.5 & 94.4 & 64.5 & 50.0 \\
 & 0-29 & 39.5 & 66.4 & 33.3 & 59.1 & 71.1 & 82.2 & 94.2 & \textbf{66.2} & 49.0 \\
 \midrule
TAR2019 & 0-1 & 43.2 & 67.1 & 25.2 & 57.3 & 73.4 & 82.0 & 95.1 & 63.2 & 52.9 \\
 & 0-2 & 45.5 & 75.4 & 25.0 & 60.7 & 75.2 & 86.8 & 95.6 & 69.7 & 59.2 \\
 & 0-4 & 47.1 & 75.6 & 27.0 & 61.8 & 76.5 & 87.5 & 95.6 & 70.1 & 60.8 \\
 & 0-9 & 48.1 & \textbf{77.6} & 27.3 & 61.0 & 76.2 & 88.6 & \textbf{96.4} & 70.7 & \textbf{63.3} \\
 & 0-14 & 48.0 & 77.5 & 28.9 & 61.6 & 75.6 & 88.0 & 96.2 & 70.9 & 63.2 \\
 & 0-19 & \textbf{50.6} & 76.5 & 29.3 & \textbf{63.6} & \textbf{76.7} & 88.3 & 95.8 & \textbf{71.2} & 60.9 \\
 & 0-24 & 48.3 & 76.8 & \textbf{29.6} & 62.0 & 76.5 & \textbf{88.8} & 96.3 & 69.9 & 62.4 \\
 & 0-29 & 48.9 & 77.1 & 27.1 & 62.3 & 76.3 & 88.6 & \textbf{96.4} & 69.6 & 62.8 \\
\bottomrule
\end{tabular}
  \caption{Comparison of different Likert scale ranges in LLM. LLM: Llama3.3-70B. Scale: 0-19. Re-ranker: monoT5 with title-only as query. %(macro recall calculation) 
  %\red{create tiny barplot of MAP for main part of paper}  
  }
  \label{tab:scales_avg}
\end{table*}

\textbf{\autoref{tab:lm_exp_avg}} shows detailed results of the comparison of dense ranking models briefly shown in \autoref{fig:dense_rankers_ablation}. 
The results confirm our choice of using monoT5 as a dense ranker with title-only as the query identified in our preliminary experiments (see \autoref{tab:dense_rankers_guo}). On both the SYNERGY and TAR2019 datasets, this configuration performs best on most metrics.

% In this experiment the setting of monoT5 as model and title as additional information performs best across most metrics.
% This experiment is an ablation experiment based on the initial experiments that were conducted in preliminary experiments (see \autoref{tab:dense_rankers_guo}).
% \chtodo{make clear that it is an ablation to initial test on Guo et al.}

\begin{table*}[ht]
\footnotesize
\centering
\setlength\tabcolsep{3pt}
\begin{tabular}{ll|rrrrrrr|rr}
\toprule
&&& & &&&&& \multicolumn{2}{c}{WSS} \\
\textbf{Dataset} & \textbf{Model} & MAP & TNR@95\% & R@1\% & R@5\% & R@10\% & R@20\% & R@50\% & @95\%& @100\% \\
\midrule
SYNERGY & BM25 (T) & 10.3 & 36.8 & 7.3 & 24.2 & 35.4 & 49.3 & 77.8 & 32.0 & 22.9 \\
 & BM25 (T+R) & 12.5 & 36.3 & 9.0 & 26.4 & 40.0 & 54.4 & 80.3 & 33.6 & 21.8 \\
 & ColBERT (T) & 10.2 & 32.4 & 7.4 & 22.9 & 34.2 & 50.7 & 80.3 & 32.9 & 20.0 \\
 & ColBERT (T+R) & 8.9 & 29.7 & 5.1 & 18.2 & 31.7 & 47.5 & 79.4 & 29.6 & 17.0 \\
 & monoBERT (T) & 16.6 & 44.6 & 13.8 & 33.6 & 47.8 & 64.1 & 87.2 & 42.6 & 30.3 \\
 & monoBERT (T+R) & 17.8 & 42.6 & \textbf{14.6} & 32.6 & 48.4 & 64.3 & 87.1 & 40.6 & 28.6 \\
 & monoT5 (T) & \textbf{18.8} & \textbf{52.3} & 14.1 & \textbf{37.5} & \textbf{54.5} & \textbf{69.1} & \textbf{91.1} & \textbf{50.0} & \textbf{35.4} \\
 & monoT5 (T+R) & 17.0 & 35.0 & 14.4 & 30.3 & 42.3 & 58.6 & 83.0 & 32.5 & 22.6 \\
 \midrule
TAR2019 & BM25 (T) & 18.1 & 38.9 & 9.5 & 30.2 & 40.5 & 57.4 & 81.4 & 36.1 & 28.4 \\
 & BM25 (T+R) & 21.1 & 44.8 & 12.6 & 34.2 & 46.2 & 60.5 & 82.5 & 40.9 & 30.4 \\
 & ColBERT (T) & 15.8 & 45.1 & 6.1 & 21.2 & 36.1 & 54.5 & 84.8 & 40.8 & 33.5 \\
 & ColBERT (T+R) & 14.5 & 43.7 & 4.7 & 24.2 & 34.2 & 52.4 & 83.7 & 39.4 & 33.7 \\
 & monoBERT (T) & 23.8 & 50.8 & 13.2 & 38.5 & 51.3 & 64.8 & 89.4 & 47.4 & 36.2 \\
 & monoBERT (T+R) & 27.4 & 54.4 & 16.9 & 40.9 & 54.3 & 71.2 & 90.8 & 49.9 & 41.0 \\
 & monoT5 (T) & \textbf{30.1} & \textbf{63.2} & 17.7 & \textbf{44.6} & \textbf{59.3} & \textbf{73.4} & \textbf{94.0} & \textbf{58.2} & \textbf{50.7} \\
 & monoT5 (T+R) & 29.6 & 59.9 & \textbf{19.0} & 42.6 & 57.1 & 72.8 & 93.2 & 56.1 & 45.8 \\
\bottomrule
\end{tabular}
  \caption{Comparison of dense ranking models. T = title of the SLR, R = research questions of the SLR used as query. All models have been trained on MS MACRO.}
  \label{tab:lm_exp_avg}
\end{table*}

% on-top
\subsection{2-shot Experiments}
\label{sec:2-shot}
\begin{table*}[htbp]
\footnotesize
\setlength{\tabcolsep}{3pt}
\begin{tabular}{ll|rrrrrrr|rr}
\toprule
&&& & &&&&& \multicolumn{2}{c}{WSS} \\
\textbf{Dataset} & \textbf{Prompt} & MAP & TNR@95\% & R@1\% & R@5\% & R@10\% & R@20\% & R@50\% & @95\%& @100\% \\
\midrule
SYNERGY & 0-shot & \textbf{40.7} & \textbf{67.0} & \textbf{34.1} & 59.3 & \textbf{72.0} & 81.9 & \textbf{94.4} & \textbf{65.2} & 49.3 \\
 & 2-shot & 38.7 & 60.5 & \textbf{34.1} & \textbf{60.6} & 71.8 & 80.7 & 92.8 & 59.5 & 44.8 \\
 & 2-shot CoT & 37.0 & 65.1 & 32.3 & 59.0 & 71.8 & \textbf{83.4} & 93.6 & 62.8 & 48.2 \\
 & 2-shot CoT (n=3) & 39.2 & 66.2 & 32.5 & 58.7 & 71.0 & 82.8 & 94.1 & 64.2 & \textbf{49.9} \\
 \midrule
TAR2019 & 0-shot & \textbf{50.6} & \textbf{76.5} & \textbf{29.3} & \textbf{63.6} & \textbf{76.7} & \textbf{88.3} & 95.8 & \textbf{71.2} & \textbf{60.9} \\
 & 2-shot & 45.4 & 72.5 & 27.3 & 57.8 & 76.0 & 85.2 & 95.7 & 68.2 & 57.1 \\
 & 2-shot CoT & 43.5 & 72.0 & 25.7 & 57.5 & 69.8 & 86.0 & \textbf{96.5} & 67.0 & 60.5 \\
 & 2-shot CoT (n=3) & 44.8 & 74.9 & 26.1 & 57.9 & 74.2 & 85.7 & 95.8 & 68.8 & 59.4 \\
\bottomrule
\end{tabular}
  \caption{Performance of 2-shot approaches. LLM: Llama3.3-70B. Scale: 0-19. Re-ranker: monoT5 with title-only as query.}
  \label{tab:prompt_avg_res}
\end{table*}

Since previous research on LLMs has shown that adding examples to the prompts can be beneficial to their performance \cite{Brown2020Language}, we test the few-shot setting for the sake of completeness: admittedly, each paper can be considered irrelevant/relevant due to various aspects of the selection criteria, i.e., it is an extremely difficult task to generate examples that are representative for all irrelevant/relevant papers.

All abstract screening datasets provide binary relevance labels, yet we work with assigning graded relevance judgments ranging from $0$ to $k$.
We use only two examples (one positive and one negative) for few-shot prompting. % because the dataset is binary, i.e., we can only give examples with extreme values. 
%Providing the LLM with even more extreme values could therefore bias the model to not use intermediate values, which is an undesirable behavior. 
In the prompt, the positive example is always followed by the negative example.
%The order of the few-shot examples is always "PN," i.e., first the positive and then the negative example. 
%For self-consistency runs, n=3 is chosen because increasing the number of repetitions would result in even higher runtime and computational cost. %\wiedholz{Which n did others choose when implementing self-consistency} \chris{really depends on the task, e.g., the orignal papers goes up to n=40, but they already say that this is quite costly and that one can try smaller n values like 5 or 10 -> but that would already be quite a lot in our application}
%Some of the evaluated prompting techniques\annetodo{CoT or others?} require few-shot examples, in particular, we provide one example of an included and one example of an excluded paper (scored $0$ and $1$, respectively).
In practice, a domain expert could provide these examples together with a formulation of the reasoning they applied.
As an approximation, we select the few-shot examples 
%(including the reasoning chain demonstrations) 
from zero-shot runs of the respective prompting technique, i.e., our default prompt or CoT prompt, for each SLR as follows.
We select the set of examples to which the zero-shot model assigns either the score $0$ (for instances labeled as irrelevant in the original dataset) or $k$ (for instances labeled as relevant in the original dataset), that exhibit the expected format, i.e., the decision of the model can be extracted using regular expressions.
We then select one positive and one negative example from these sets randomly and remove them from the evaluation sets.
%\annetodo{just from the few-shot runs? how many are you using? there are no training splits, right? -- das hier ist im Prinzip ein wenig problematisch, die hätten dann eigentlich komplett zur Seite gelegt gehört  - kannst du die Scores entsprechend nochmal berechnen? zur Not nach der Submission, ich gehe auch davon aus, dass die Scores sich eigentlich nicht ändern}
%We test the few-shot setting for the sake of completeness: admittedly, each paper can be considered irrelevant/relevant due to various aspects of the selection criteria, i.e., it is an extremely difficult task to generate examples that are representative for all irrelevant/relevant papers.

In \textbf{\autoref{tab:prompt_avg_res}}, we compare the performance of 0-shot prompting with 2-shot, 2-shot CoT, and 2-shot CoT with self-consistency.  %we compare the performance of 0-shot with 2-shot and 2-shot CoT prompting. 
The results suggest that adding examples does not improve the performance of LGAR, and in fact it is harmful. 
The reason for this could be that the randomly chosen examples are not representative of all irrelevant/relevant papers. 
Furthermore, since the examples are generated by the LLM, they could also contain reasoning errors. 
This could be improved by having domain experts generate the examples. 
However, the challenge of finding good representatives for an SLR remains. 
Therefore, we did not pursue the use of few-shot examples.

\subsection{Hardware and Experiment Statistics}
We ran our experiments on Nvidia A40 (40 GB) and Nvidia A100 (80 GB) GPUs, using up to 8 GPUs in parallel. The total amount of GPU hours was 7345h on A40 and about 3664h on A100. %\chtodo{have no statistic for Licca -> leave it out or estimate it?}

% statistics of FAU Cluster:
% \begin{itemize}
%     \item Nvidia A40 (40GB): 7345h
%     \item Nvidia A100 (80GB): 3664h (1480h for the paper, rest is of bachelor thesis)
% \end{itemize}
% however, no statistics for Licca; up to 8 GPUs in parallel

\subsection{Use of AI Assistants}
Copilot and ChatGPT were used only as coding assistants. This was especially useful for debugging existing code and for understanding third-party code. We have not used AI assistants for writing this paper. %\chtodo{passt das so oder fragt man sich dann, was foreign code ist?} %\chtodo{muss man da mehr dazu schreiben?}

\end{document}